\documentclass[sigconf,nonacm]{acmart}

\AtBeginDocument{%
  }

\renewcommand\footnotetextcopyrightpermission[1]{}
\usepackage{amsmath,amssymb,mathtools,amsthm,bm}

\newcommand{\xb}{\mathbf{x}}
\newcommand{\zb}{\mathbf{z}}
\newcommand{\bdelta}{\bm{\delta}}
\newcommand{\btheta}{\bm{\theta}}
\newcommand{\norm}[1]{\left\|#1\right\|}

\newcommand{\header}[1]{\noindent\textbf{#1}}

\usepackage{hyperref,url,multirow,inputenc,fontenc,nicefrac,float,microtype}
\usepackage{graphicx,subcaption,xcolor,booktabs,lipsum,xspace,enumitem}
\usepackage{amsfonts,colortbl,algorithm,algpseudocode,doi}
\usepackage[most]{tcolorbox}
\usepackage{xurl}
\usepackage[capitalize,noabbrev]{cleveref}
\usepackage{array,pifont,wrapfig,natbib,multicol}
\usepackage{balance}

\definecolor{lightblue}{rgb}{0.68, 0.85, 0.9}
\definecolor{lightgreen}{rgb}{0.68, 0.9, 0.68}
\definecolor{lightred}{rgb}{0.9, 0.68, 0.68}
\definecolor{lightpurple}{rgb}{0.9, 0.68, 0.9}
\definecolor{lightyellow}{rgb}{0.9, 0.9, 0.68}
\definecolor{darkred}{rgb}{0.6,0,0}
\definecolor{darkblue}{rgb}{0,0,0.7}
\definecolor{darkgreen}{rgb}{0,0.5,0}

\newcommand{\secondrunner}[1]{\cellcolor{gray!20}{#1}}

\newcommand{\cmarkcolor}{\textcolor{darkgreen}{\ding{51}}}
\newcommand{\xmarkcolor}{\textcolor{darkred}{\ding{55}}}

\newcommand{\blueUpdate}[1]{#1}

\setcitestyle{numbers,square}

\hypersetup{
    colorlinks=false,
    linkcolor=black,
    filecolor=black,
    urlcolor=black,
    citecolor=black
}

\title{Rethinking and Red-Teaming Protective Perturbation in Personalized Diffusion Models}

\author{Yixin Liu}
\authornote{Correspondence to: yila22@lehigh.edu, lis221@lehigh.edu}
\affiliation{%
  \institution{Lehigh University}
  \department{Computer Science and Engineering}
  \city{Bethlehem}
  \state{PA}
  \country{USA}
}

\author{Ruoxi Chen}
\authornote{Work done while a visiting scholar at Lehigh University.}
\affiliation{%
  \institution{Lehigh University}
  \department{Computer Science and Engineering}
  \city{Bethlehem}
  \state{PA}
  \country{USA}
}

\author{Xun Chen}
\affiliation{%
  \institution{Independent Researcher}
  \city{Fremont}
  \state{California}
  \country{USA}
}

\author{Lichao Sun}
\affiliation{%
  \institution{Lehigh University}
  \department{Computer Science and Engineering}
  \city{Bethlehem}
  \state{PA}
  \country{USA}
}


\renewcommand{\shortauthors}{Yixin Liu, Ruoxi Chen, Xun Chen, and Lichao Sun}

\begin{document}

\begin{abstract}

    Personalized diffusion models (PDMs) have become prominent for adapting pre-trained text-to-image models to generate images of specific subjects using minimal training data. However, PDMs are susceptible to minor adversarial perturbations, leading to significant degradation when fine-tuned on corrupted datasets. These vulnerabilities are exploited to create protective perturbations that prevent unauthorized image generation. Existing purification methods attempt to red-team the protective perturbation to break the protection but often over-purify images, resulting in information loss. In this work, we conduct an in-depth analysis of the fine-tuning process of PDMs through the lens of shortcut learning. We hypothesize and empirically demonstrate that adversarial perturbations induce a latent-space misalignment between images and their text prompts in the CLIP embedding space. This misalignment causes the model to erroneously associate noisy patterns with unique identifiers during fine-tuning, resulting in poor generalization. Based on these insights, we propose a systematic red-teaming framework that includes data purification and contrastive decoupling learning. We first employ off-the-shelf image restoration techniques to realign images with their original semantic content in latent space. Then, we introduce contrastive decoupling learning with noise tokens to decouple the learning of personalized concepts from spurious noise patterns. Our study not only uncovers shortcut learning vulnerabilities in PDMs but also provides a thorough evaluation framework for developing stronger protection. Our extensive evaluation demonstrates its advantages over existing purification methods and its robustness against adaptive perturbations. Code is available at \url{https://github.com/liuyixin-louis/DiffShortcut}.
  \end{abstract}

\maketitle

\begin{figure}[t!]
    \centering
    \includegraphics[width=\linewidth]{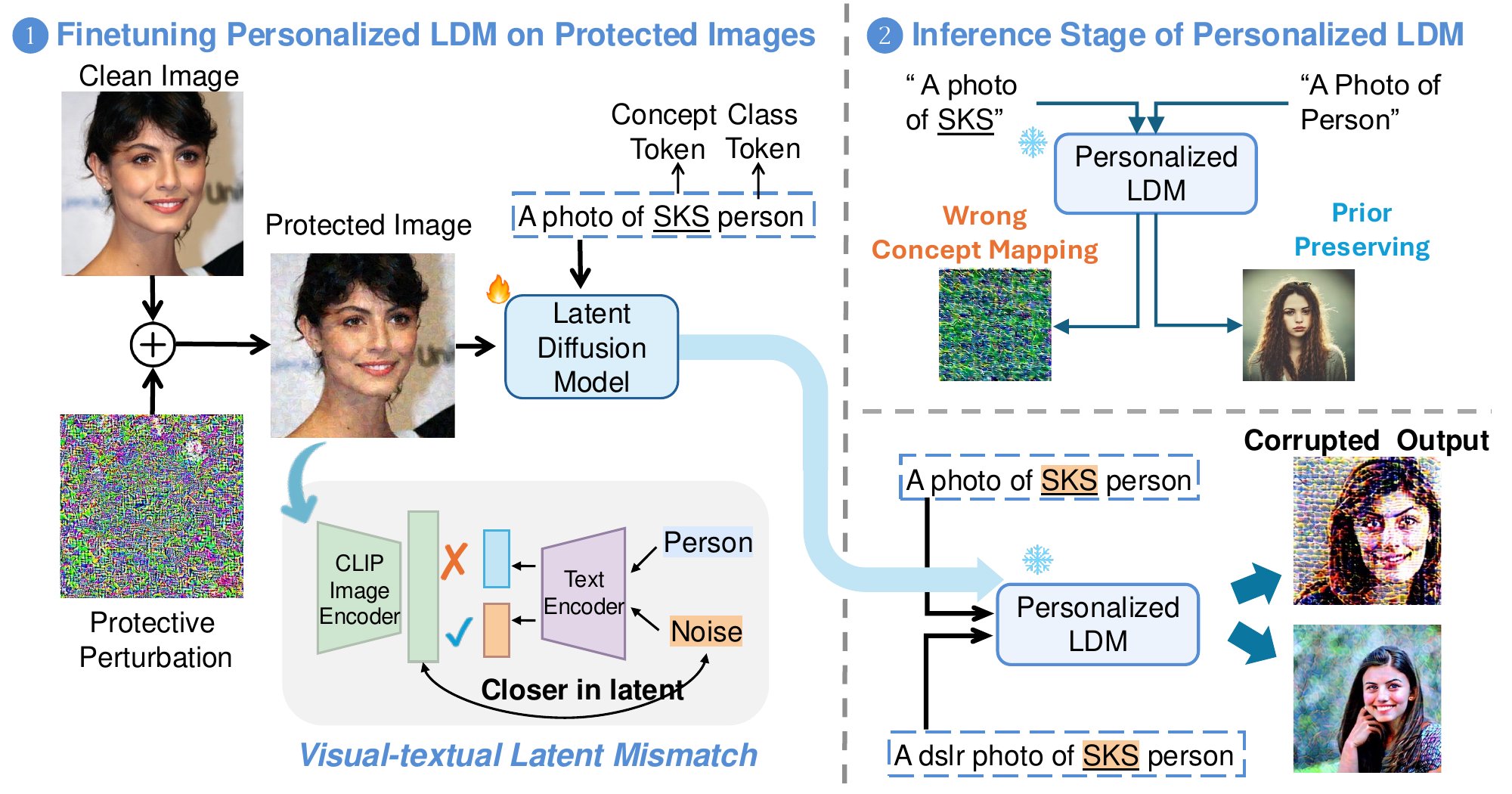}
    \caption{We observe that protective perturbation for personalized diffusion models creates a latent mismatch in the image-prompt pair. Fine-tuning on such perturbed data tricks the models, learning the wrong concept mapping. Thus, model generations suffer from severe degradation in quality. }
    \label{fig:teaser}
\end{figure}

\section{Introduction}
\label{sec. intro}

% \rewrite{
The rapid advancements in text-to-image diffusion models, such as DALL-E 2~\citep{ramesh2022hierarchical}, Stable Diffusion~\citep{Rombach_2022_CVPR}, and MidJourney~\citep{midjourney_website}, have revolutionized the field of image generation. These models can generate highly realistic and diverse images based on textual descriptions, enabling a wide range of applications in creative industries, entertainment, and beyond. 
However, the ability to fine-tune these models on small, personal datasets raises significant concerns. This includes potential misuse for generating misleading content targeting individuals~\citep{van2023anti,salman2023raising} and threatening artists' livelihoods by mimicking their unique styles without compensation~\citep{shan2023glaze}. As shown in Fig.~\ref{fig:teaser}, we observe that protective perturbation can cause a latent mismatch between the image and prompt, leading to degraded generation quality.
% }

To address these issues, several \emph{protective perturbation} methods have been proposed to protect user images from unauthorized personalized synthesis~\citep{deng2024survey,wang2024replication}. These methods aim to make images resistant to AI-based manipulation by crafting adversarial perturbations~\citep{salman2023raising,liang2023adversarial}, applying subtle style-transfer cloaks~\citep{shan2023glaze}, or crafting misleading perturbations that cause the model to overfit \citep{Liu_2024_CVPR}. The model trained on perturbed data will generate images that are poor in quality, and thus, the unauthorized fine-tuning fails. For standard off-the-shelf models, the mechanism is relatively well understood, often linked to the adversarial vulnerability of neural networks~\citep{ilyas2019adversarial} or the latent space characteristics of VAEs~\citep{kingma2013auto,guo2024smooth,xue2023toward}. However, the underlying mechanism for how these perturbations disrupt the fine-tuning process of \textit{personalized} diffusion models remains unexplored.

% Moreover, purification studies are also purposed to further break those protections. 
\blueUpdate{Moreover, to systematically examine the practical performance of existing protection methods in the wild, purification studies \citep{cao2024impress,zhao2024can} have been proposed with more advanced data purification processes to further re-evaluate and red-team these protection methods.}
As demonstrated in \citet{van2023anti}, most of the protection methods lack resilience against simple purification like Gaussian smoothing. \blueUpdate{However, these traditional transformations also come with severe data quality degradation after purification.} 
% However, this simple rule-based purification causes severe data quality degradation. 
% As demonstrated in \citet{cao2024impress,zhao2024can}, \blueUpdate{compared to traditional rule-based purifications like JPEG compression \citep{van2023anti}}, 
\blueUpdate{Compared to these deterministic purifications,} \blueUpdate{diffusion-based} purification shows a stronger capacity to denoise the images and yield high-quality output by leveraging the distribution modeling ability of diffusion models. IMPRESS~\citep{cao2024impress} observes that clean images show better reconstruction consistency. It therefore optimizes protected images by imposing a visual LPIPS similarity constraint~\citep{lpips} to enforce this consistency. Despite its effectiveness, \blueUpdate{IMPRESS} \blueUpdate{is inefficient and} requires a tremendous amount of time due to the iterative nature of the proposed optimization. On the other line, 
% DiffPure \citep{nie2022diffusion} proposes to leverage off-the-shelf pixel-space diffusion models to conduct an SDEdit process \citep{meng2021sdedit} that converts the perturbed images into a certain noisy state with a diffusion forward process and then denoises in the reverse process. 
GrIDPure~\citep{zhao2024can} uses pixel-space diffusion models for denoising. It employs an SDEdit-like process~\citep{meng2021sdedit,nie2022diffusion}, which first adds slight noise to an image before denoising it via a reverse process. To further improve visual consistency, GrIDPure also divides images into smaller grids and applies a small-step diffusion process.
% further divides the images into smaller grids and employs small-step DiffPure that yields better visual consistency. 
However, \blueUpdate{GrIDPure} still yields \blueUpdate{\textit{unfaithful}} content that causes substantial identity shift due to the generative nature of the diffusion model. \textit{How to design an effective, efficient, and faithful purification approach is still an open question. }

% Given the above two limitations of existing works, i
% In this paper, we first take a closer look at the underlying mechanism of the effectiveness of protective protection. On the representation side, we found that the latent of the perturbed images are largely shifted from their original semantic concepts. As a consequence, on the model side, we discover that the PDMs turn to falsely link the personalized concept token with the injected noisy perturbation instead of learning the clean identity concept shown in the image. Based on these findings, we uncovers the shortcut-learning vulnerability of personalized diffusion when fine-tuning the model on perturbed data. 
\blueUpdate{To provide a \blueUpdate{mechanistic diagnosis} of why protections succeed, we analyze the fine-tuning process of PDMs through the lens of \emph{causal analysis} and \emph{shortcut learning} \citep{geirhos2020shortcut}. We first build the underlying causal graph of learning on protected images, where we find that protective perturbations manipulate the learning process by reinforcing the shortcut path from personalized identifier to injected noise. Furthermore, we find that existing effective protective perturbations introduce a latent-space misalignment between images and the textual prompts, where the perturbed images largely deviate from their original semantic concepts. 
This misalignment triggers the model to learn a shortcut connection between the identifier and more high-frequency and easy-to-learn noise patterns. 
% To support our hypothesis, we conduct latent space visualization and concept prompting interpretation, where we found protected images are largely deviated from their original semantic concepts. 
}
% To validate our hypothesis, we conduct comprehensive experiments demonstrating that \blueUpdate{protected} images significantly deviate from their semantic concepts in the latent space. \blueUpdate{And thus, }fine-tuning on such mismatched image-prompt pairs leads the model to learn spurious correlations toward those more available noise patterns.
% . This misalignment causes the model to erroneously associate the injected noise patterns with the unique identifier tokens during fine-tuning, resulting in degraded performance. To validate our hypothesis, we conduct comprehensive experiments demonstrating that \blueUpdate{protected} images significantly deviate from their semantic concepts in the latent space. \blueUpdate{And thus, }fine-tuning on such mismatched image-prompt pairs leads the model to learn spurious correlations toward those more available noise patterns.

% Motivated by this empirical understanding, we present a systematic purification and disentangled training approach to empower robust personalized diffusion models upon perturbed data. 
Based on these insights, we propose a systematic \blueUpdate{red-teaming} framework \blueUpdate{motivated by causal intervention} to empower robust PDMs against \blueUpdate{protective} perturbations.
Our framework offers three main advantages over existing methods, which are often limited to image-level purification:
\begin{itemize}
    \item \textbf{Efficiency and Faithfulness:} We use efficient, one-shot super-resolution and restoration models to purify images, converting low-quality inputs into high-quality results.
    \item \textbf{Robustness:} Our contrastive decoupling learning works independently and provides strong robustness against adaptive attacks targeting our pipeline.
    \item \textbf{System-Level Red-Teaming:} We propose a comprehensive, system-level strategy that covers data purification, model training, and sampling, offering a more thorough evaluation of protection effectiveness.
\end{itemize}
We summarize our contributions as follows:
\begin{itemize} [nolistsep, leftmargin=*]

    % \item We empirically explain the underlying mechanism of the success of the recent protective perturbations approach through the lens of shortcut learning. 
    % We find that effective perturbation causes \textit{latent-space image-prompt mismatch}, where the perturbed images no longer align with their original prompt pair in latent space and thus training on such contradicted pairs tricks the model into learning false correlations. We validate this through comprehensive latent visualization and concept interpretation experiments. 
    
    % \item  Motivated by this understanding, we propose a systematic defensive framework that empowers robust personalized diffusion models from input image-text pair, model training, and the sampling process. Specifically, we first leverage off-the-shelf image restoration models to realign the image with its prompt's semantic meaning; on model training, we propose contrastive decouple learning with the negative token to disentangle the learning of personalized concepts from spurious noisy patterns; on the sampling, we propose quality-enhanced sampling strategy with negative prompting. 
    % % As a result, our method is more effective, efficient, and faithful. 
    
    % \item We empirically demonstrate the effectiveness, efficiency, and faithfulness of our framework on the facial dataset VGGFace2. Moreover, under challenging adaptive attack settings, we demonstrate our framework is more robust than other variants. 
    
    \item We provide the first mechanistic diagnosis revealing that \blueUpdate{\blueUpdate{protective} perturbations work by exploiting the shortcut learning in PDMs with latent-space image-prompt misalignment from causal analysis.}
    \item We propose a systematic \blueUpdate{red-teaming} framework based on causal analysis that effectively mitigates these vulnerabilities through data purification and contrastive decoupling learning and sampling.
    \item We demonstrate the effectiveness, efficiency, and faithfulness of our approach through extensive experiments across 7 protections, showing significant improvements over existing methods. Our study provides a more \blueUpdate{systematic} evaluation framework for future research on protective perturbations.
\end{itemize}

% \vspace{-5pt}
\section{Related Works}
\label{sec. related work}
% \vspace{-5pt}

% \textbf{Latent Diffusion Models and Customized Adaption.} 
% Generative Models (GMs) aim to synthesize samples from a data distribution given a set of training examples. Diffusion Probabilistic Models \citep{ho2020denoising} are now the dominant GMs with various applications such as text-to-image synthesis \citep{Rombach_2022_CVPR}, image editing \citep{choi2023customedit,kim2022diffusionclip,shi2023dragdiffusion}, and image inpainting \citep{bar2022visual}. To improve the training efficiency, Latent Diffusion Models (LDMs) \citep{Rombach_2022_CVPR} are proposed to conduct a more efficient diffusion process in a low-dimensional latent space with a pair of pretrained image encoder and decoder (e.g., VAE \citep{kingma2013auto}). Despite the general capability, model users often wish to synthesize specific concepts from their own personal lives. To meet that, personalized text-to-image generation \citep{kumari2022customdiffusion, ruiz2023dreambooth,ruiz2023hyperdreambooth,raj2023dreambooth3d} is proposed to learn the concepts using a small set of reference images. Among them, DreamBooth~\citep{ruiz2023dreambooth} is one of the mainstream approaches that involve full fine-tuning of parameters, yielding more superior generation performance compared to those only fine-tune partial parameter \citep{hu2021lora} or a pseudo-word vector \citep{gal2022image}. Besides, it uses a rarely-used token to link the concept and incorporate prior-preserving loss with class-oriented images, enabling a high-quality and diverse generation of subjects.

\textbf{Data Poisoning as Protection against Unauthorized Training with LDMs.}
Latent Diffusion Models (LDMs) \citep{Rombach_2022_CVPR} have become dominant in various generative tasks, including text-to-image synthesis. To meet the demand for personalized generation, methods like DreamBooth~\citep{ruiz2023dreambooth} have been proposed, which fine-tune LDMs using a small set of reference images to learn specific concepts. However, these advancements have raised concerns about potential misuse, such as generating misleading content targeting individuals~\citep{van2023anti,salman2023raising} and threatening the livelihood of professional artists through style mimicking~\citep{shan2023glaze}. To address these issues, several data-poisoning-based methods have been proposed to protect user images from unauthorized personalized synthesis by injecting adversarial perturbations through minimizing adversarial target loss in image encoder or UNet denoiser~\citep{salman2023raising}, or denoising-loss maximization~\citep{liang2023adversarial, van2023anti, Liu_2024_CVPR} or in opposite direction, denoising-loss minimization \citep{xue2023toward}, or cross-attention loss maximization~\citep{xu2024perturbing}. Beyond personalized generation, similar protection paradigms have been extended to medical imaging~\citep{liu2023securing,sun2024medical}, image editing~\citep{chen2024editshield}, and classifier training~\citep{liu2024stable}. 
Despite its effectiveness, a mechanistic diagnosis of why protections succeed against diffusion model fine-tuning has not yet been provided. To the best of our knowledge, \citet{zhao2024can} is the only work that attempts to investigate the underlying mechanism. However, it is limited to the vulnerability of the text encoder. \textit{In this work, we provide a more comprehensive explanation from the perspective of latent mismatch and shortcut learning. }

\textbf{Data Purification that Further Breaks Protection.} 
Despite promising protection performance, studies \citep{van2023anti,an2024rethinking,Liu_2024_CVPR} suggest that these perturbations without advanced transformation loss \citep{athalye2018synthesizing} are brittle and can be easily removed under simple rule-based transformations.
% However, adaptive protection with EoT \citep{athalye2018synthesizing} indicates that protection can further bypass these simple rule-based transformations. 
Among all types of transformation, state-of-the-art adversarial purification leverages diffusion models as purifiers to perturb images back to their clean distributions. 
In the classification scenario, DiffPure \citep{nie2022diffusion} is a mainstream approach for adversarial purification by applying SDEdit on the poison with an off-the-shelf diffusion model. For purification against protective perturbation, GrIDPure \citep{zhao2024can} further adapts iterative DiffPure with small steps on multi-grid split image to preserve the original resolution and structure. However, due to their generative nature, these SDEdit-based purifications have limitations in yielding unfaithful content~\citep{hu2024improving}, where the purified images fail to preserve the original identity. Observing the perceptible inconsistency between the perturbed images and the diffusion-reconstructed ones, IMPRESS \citep{cao2024impress} conducts the purification via minimizing the consistency loss with constraints on the maximum LPIPS-based similarity change on pixel space. While it manages to preserve similarity, IMPRESS suffers from the inefficiency issue due to its iterative process and is ineffective under stronger protections like \citet{Liu_2024_CVPR,mi2024visual}. 

\textbf{Shortcut Learning and Causal Analysis.}
Shortcut learning occurs when models exploit spurious correlations in training data, leading to poor generalization \citep{geirhos2020shortcut}. The causal analysis provides a framework for addressing this by modeling cause-effect relationships \citep{pearl2009causal,scholkopf2021toward}. It helps identify true causal factors, distinguishing them from spurious correlations. In PDMs, \blueUpdate{protective} perturbations can introduce spurious correlations between noise patterns and identifiers during fine-tuning. \textit{Our work explores how to restore correct causal relationships when learning PDMs on perturbed data, which is under-explored in existing works.}

\section{Preliminary}
\label{sec:preliminary}

\textbf{Personalized Latent Diffusion Models (LDMs) via DreamBooth Fine-tuning.}
LDMs~\citep{Rombach_2022_CVPR} are generative models that perform diffusion processes in a lower-dimensional latent space, enhancing training and inference efficiency compared to pixel-space diffusion models~\citep{ho2020denoising}. By conditioning on additional embeddings such as text prompts, LDMs can generate or edit images guided by these prompts. Specifically, an image encoder $\mathcal{E}$ maps an image $\xb_0$ to a latent representation $\zb_0 = \mathcal{E}(\xb_0)$. A text encoder $\tau_{\theta}$ produces a text embedding $\bm{c} = \tau_{\theta}(c)$ for a given prompt $c$. The model trains a conditional noise estimator $\bm{\epsilon}_{\theta}$, typically a UNet~\citep{ronneberger2015u}, to predict the Gaussian noise added at each timestep $t$, using the loss:
\begin{equation}
\mathcal{L}_{\text{denoise}}(\xb_0, \bm{c}; \theta) = \mathbb{E}_{\zb_0 \sim \mathcal{E}(\xb_0), \bm{\epsilon}, t }\Big[ \Vert \bm{\epsilon} - \bm{\epsilon}_{\theta}(\zb_0,t, \bm{c}) \Vert_{2}^{2}\Big].
\label{eq:denoise_loss}
\end{equation}
During inference, the model starts from random noise $\zb_T \sim \mathcal{N} (0, \textbf{I})$ and iteratively denoises it to obtain a latent $\tilde{\zb}_0$, which is then decoded to generate the image $\tilde{\xb}_0 = \mathcal{D}(\tilde{\zb}_0)$. DreamBooth~\citep{ruiz2023dreambooth} fine-tunes a pre-trained LDM to generate images of specific concepts by introducing a unique identifier that links subject concepts and employing a class-specific prior-preserving loss to mitigate overfitting and language drift. The fine-tuning utilizes an instance dataset $\mathcal{D}_{\bm{x}_0}=\left\{ \left( \bm{x}_0^{i},\bm{c}^{\blueUpdate{\mathcal{V}^*}} \right) \right\}_i$, and a class dataset $\mathcal{D}_{\bar{\bm{x}}_0}=\left\{ \left( \bar{\bm{x}}_0^{i},\blueUpdate{\bar{\bm{c}}} \right) \right\}_i$, where $\bm{x}_0$ are subject images and $\bar{\bm{x}}_0$ are class images. The class-specific prompt \blueUpdate{$\bar{\bm{c}}$} is set as \textit{``a photo of a [class noun]''}, and the instance prompt \blueUpdate{$\bm{c}^{\blueUpdate{\mathcal{V}^*}}$} is \textit{``a photo of \blueUpdate{$\mathcal{V}^*$} [class noun]''}, where \blueUpdate{ $\mathcal{V}^*$} specifies the subject and ``[class noun]'' denotes the object category (e.g., ``person''). 
\blueUpdate{The instance dataset contains the subject-specific images we want the model to learn, while the class dataset contains diverse images from the same category to prevent language drift.}
The fine-tuning process \blueUpdate{on these two datasets} optimizes a weighted sum of the instance denoising loss and the prior-preservation loss: 
% to achieve personalization while preserving the model's understanding of the general class concept. 
\begin{equation}
    \label{eq:DB_loss}
\mathcal{L}_{db}(\xb_0,\bm{c}^{\blueUpdate{\mathcal{V}^*}},\bar{\bm{x}}_0,\blueUpdate{\bar{\bm{c}}};\bm{\theta} ) = \mathcal{L}_{\mathrm{denoise}}\left( \xb_0,\bm{c}^{\blueUpdate{\mathcal{V}^*}} \right) + \lambda \mathcal{L}_{\mathrm{denoise}}\left( \bar{\bm{x}}_0,\blueUpdate{\bar{\bm{c}}} \right),
\end{equation}
where $\lambda$ balances the two terms. With approximately 1k training steps and around four subject images, DreamBooth can generate vivid, personalized subject images \citep{von-platen-etal-2022-diffusers}.
\textbf{Protective Perturbation against Personalized LDMs. } 
Recent studies suggest that minor adversarial perturbation to clean images can significantly disturb the learning of customized diffusion and also prevent image editing with an off-the-shelf diffusion model by greatly degrading the quality of the generated image. Existing protective perturbation can be classified into two categories: perturbation crafted with fixed diffusion models and perturbation crafted with noise-model alternative updating. In this paper, we focus on the second category since they are more effective in the fine-tuning setting. The general framework of these protective perturbation methods is to craft noise that maximizes an adversarial loss $\mathcal{L}_{adv}$ (often instantiated as the denoising loss $\mathcal{L}_{\text{denoise}}$) and alternately update the noise generator surrogates $\bm{\theta}^\prime$ (which can be a single model \citep{van2023anti} or an ensemble \citep{Liu_2024_CVPR}) or the attention modules \citep{xu2024perturbing}. Formally, at the $j$-th alternative step, the noise surrogate $\bm{\theta} ^{\prime}_j$ and perturbation $\bdelta ^{(j)}$ are updated via solving, 
\begin{equation}
    % \small
    \begin{aligned}
    {\bm{\theta} ^{\prime}}_j &\gets \underset{{\btheta ^{\prime}}}{\mathrm{arg}\min}\sum_x{\mathcal{L} _{db}}\left( \xb+\bdelta ^{(j-1)},\bm{c}^{\mathcal{V}^*},\bar{\bm{x}},\blueUpdate{\bar{\bm{c}}};{\btheta ^{\prime}}_{j-1} \right); \\
    \bdelta ^{(j)} &\gets \underset{\norm{\bdelta ^{(j-1)}}_{\infty}\le r}{\mathrm{arg}\max}\mathcal{L} _{adv\,\,}\left( \xb+\bdelta ^{(j-1)},\blueUpdate{\bar{\bm{c}}};{\btheta ^{\prime}}_j \right) .
    \end{aligned}
% \vspace{-2pt}
\end{equation}
To solve this, standard Gradient Descent is performed on the model parameter while the images are updated via Projected Gradient Descent (PGD) \citep{madry2018towards} to satisfy the $\ell_{\infty}$-ball perturbation budget constraint \blueUpdate{with radius $r$}, 
\begin{equation}
    \begin{aligned}
        \btheta _i &\gets \btheta _{i-1}-\beta \nabla _{\btheta _{i-1}}\mathcal{L} _{db}; \\
        \xb^{k+1} &\gets \Pi_{B_{\infty}\left( \xb^0,\blueUpdate{r} \right)}\left[ \xb^k+\eta \blueUpdate{\cdot }\mathrm{sign} \nabla _{\xb^k}\mathcal{L} _{adv}\left( \xb^k \right) \right],  
    \end{aligned}
\end{equation}
where $\Pi_{B_\infty(\xb^0, r)}(\cdot)$ is a projection operator on the $\ell_\infty$ ball that ensures $\xb^k \in B_\infty(\xb^0, r)=\left\{\xb^\prime: \Vert \xb' - \xb^0\Vert_\infty \leq r \right\} $, $\eta$ denotes the PGD step size, and the total number of PGD steps is $K$. 

% \textbf{Causal Analysis and Structural Causal Model.} Causal analysis provides a framework for understanding and modeling cause-and-effect relationships between variables \citep{pearl2009causal}, which is crucial for identifying spurious correlations and mitigating shortcut learning issues \cite{geirhos2020shortcut}. A Structural Causal Model (SCM) is a mathematical representation that uses structural equations and a directed acyclic graph (DAG) to model causal relationships among variables. An SCM consists of a set of endogenous variables $\mathbf{V}$ (variables determined within the system), a set of exogenous variables $\mathbf{U}$ (external variables influencing the system), and a set of structural equations ${f_i}$. Each endogenous variable $V_i \in \mathbf{V}$ is defined as a function of its parent variables $\text{Pa}(V_i)$ and an exogenous disturbance $U_i$: $V_i = f_i(\text{Pa}(V_i), U_i)$. Causal analysis can help mitigate shortcut learning by identifying and intervening on spurious correlations, allowing models to focus on true causal relationships rather than superficial statistical patterns. Please refer to \citet{pearl2009causal} for more details. 

\textbf{Causal Analysis and Structural Causal Model.} Causal analysis models cause-and-effect relationships between variables \citep{pearl2009causal}, helping identify spurious correlations and mitigate shortcut learning \cite{geirhos2020shortcut}. A Structural Causal Model (SCM) uses structural equations and a directed acyclic graph to represent causal relationships. It comprises endogenous variables $\mathbf{V}$, exogenous variables $\mathbf{U}$, and structural equations ${f_i}$, where each $V_i \in \mathbf{V}$ is defined as $V_i = f_i(\text{Pa}(V_i), U_i)$. By intervening on spurious correlations, causal analysis helps models focus on true causal relationships rather than superficial patterns. For more details, see \citet{pearl2009causal,geirhos2020shortcut}.

\textbf{Threat Model.} In this work, we consider a threat model where an adversary aims to perform unauthorized fine-tuning of a personalized diffusion model (PDM) using publicly available images of a user (e.g., from social media). These images are protected by adversarial perturbations designed to disrupt the fine-tuning process. The adversary's goal is to bypass these protections and successfully fine-tune a PDM on the perturbed images to generate new, high-quality images of the user without their consent. We assume a white-box adversary who knows the defense algorithm but not the clean originals. The adversary can access the perturbed images but cannot obtain the original, unprotected images. This scenario reflects a realistic setting where individuals seek to protect their online photos from being misused in generative models.

\section{Methodology}
\label{sec:method}

% \begin{figure}
%     \centering
%     \includegraphics[width=.9\linewidth]{figures/merge-three.png}
%     \caption{Caption}
%     \label{fig:3.1interpretation}
% \end{figure}

\subsection{
\blueUpdate{Protective} Perturbation Causes Latent-space Image-Prompt Mismatch
}

\begin{figure}[!t]
    \centering
    \includegraphics[width=\linewidth]{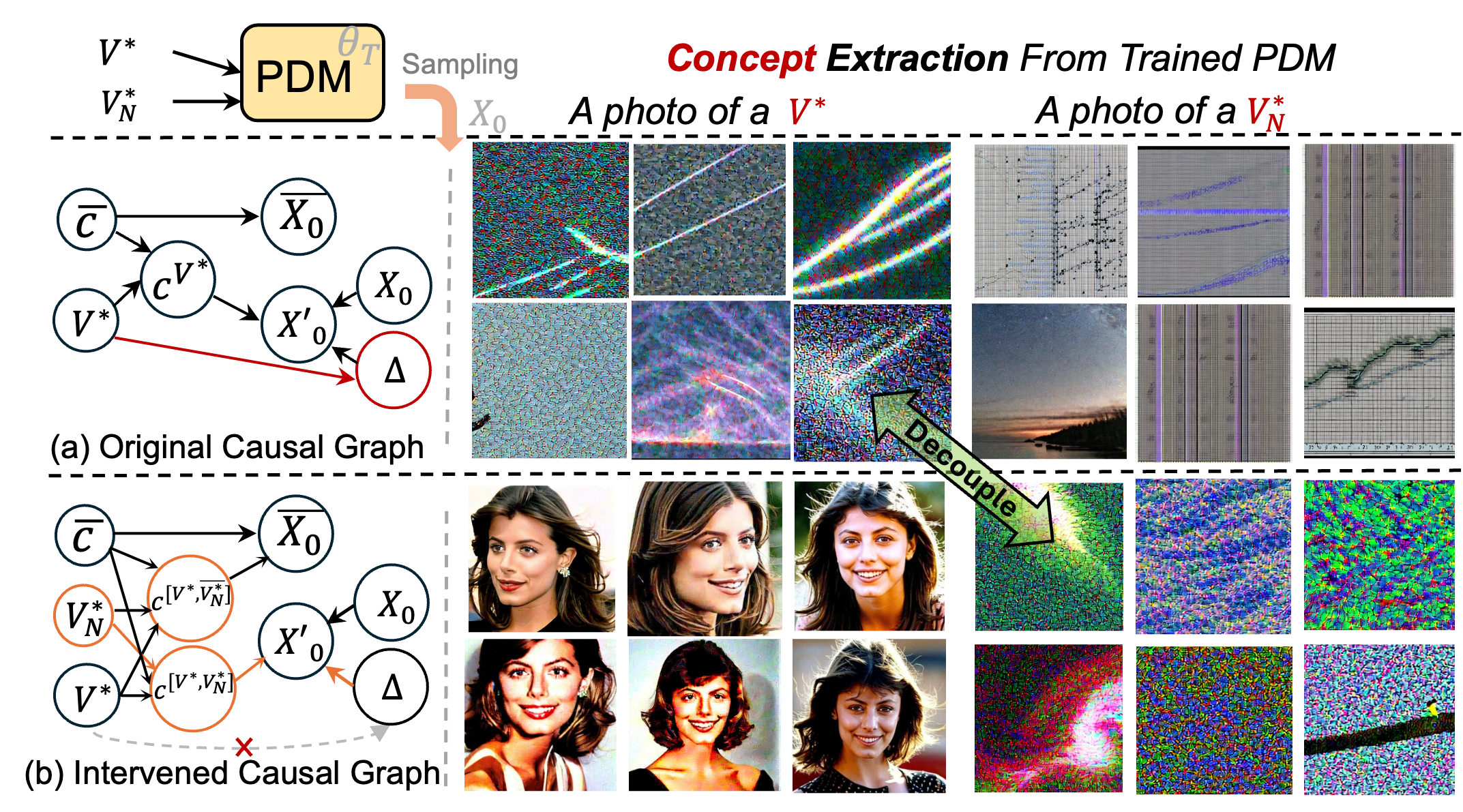}
    \caption{
        Causal view of shortcut learning. (a) Protective perturbations create a spurious shortcut (red arrow) between the identifier $\mathcal{V}^*$ and noise $\Delta$. (b) Our CDL introduces a noise token $\mathcal{V}_N^*$ (orange arrows) to decouple noise from $\mathcal{V}^*$, enabling clean concept learning.
    }
    % Original lengthy caption:
    % Causal view of shortcut learning and the effect of our Contrastive Decoupling Learning (CDL). (a) The original causal graph shows that protective perturbations create a spurious shortcut (red arrow) between the personalised identifier ($\mathcal{V}^*$) and the noise pattern ($\Delta$). This causes the model to associate the identifier with noise. (b) Our intervened causal graph where CDL introduces a new token ($\mathcal{V}_N^*$) to explicitly model the noise (orange arrows), decoupling it from $\mathcal{V}^*$. The visualisations on the right demonstrate this: with concept extraction, we see that using CDL successfully decouples the original noise pattern (spuriously linked to $\mathcal{V}^*$) and re-associates it with the new noise token $\mathcal{V}_N^*$.
    \label{fig:causal-graph}
\end{figure}

We first derive the formulation of learning personalized diffusion models on perturbed data. For the case of data poisoning, the instance data is perturbed by some adversarial noise $\bdelta$, and the personalized diffusion models optimize the following loss,
% % \vspace{-6pt}
\begin{equation}
    \label{eq. adv loss}
\mathcal{L} _{db}^{adv}(\xb_0,\bm{c}^{\mathcal{V}^*},\bar{\bm{x}}_0,\blueUpdate{\bar{\bm{c}}};\bm{\theta} )=\mathcal{L} _{\mathrm{denoise}}\left( \xb_0+\bdelta,\bm{c}^{\mathcal{V}^*} \right) +\lambda \mathcal{L} _{\mathrm{denoise}}\left( \bar{\bm{x}}_0,\blueUpdate{\bar{\bm{c}}} \right).
\end{equation}

% \begin{wrapfigure}{r}{0.33\textwidth}
%   \centering
%   \includegraphics[width=0.33\textwidth]{figures/bar_plot_with_error_bars.pdf}
%   \caption{Zero-shot classification with CLIP-based classifier.}
%   \label{fig:clip-zero}
% \end{wrapfigure}

% \begin{figure}[t]
%     \begin{minipage}[t]{0.63\textwidth}
%         \centering
%         \includegraphics[width=\linewidth]{figures/bar_plot_with_error_bars.pdf}
%         \caption{Zero-shot classification with CLIP-based classifier.}
%         \label{fig:clip-zero}
%     \end{minipage}
%     \hfill
%     \begin{minipage}[t]{0.36\textwidth}
%         \centering
%         \includegraphics[width=\linewidth]{clip-vis.png}
%         \caption{\textbf{2D visualization of CLIP-based latent for clean and perturbed cases.}}
%         \label{fig:clip-vis}
%     \end{minipage}
% \end{figure}

Based on the adversarial loss in Eq.~\ref{eq. adv loss}, we build the underlying causal graph to represent the learning process. As shown in Fig.~\ref{fig:causal-graph}(a), without intervention, protective perturbations create a spurious shortcut between the personalized identifier ($\mathcal{V}^*$) and the noise pattern ($\Delta$). This causes the model to incorrectly associate the identifier with noise. The visualization confirms this: when prompted with ``a photo of a $\mathcal{V}^*$ Person,'' the model trained on perturbed data generates a noisy, distorted image, failing to learn the true personalized concept.

In contrast, Fig.~\ref{fig:causal-graph}(b) demonstrates the effectiveness of our proposed Contrastive Decoupling Learning (CDL). By introducing a new noise token ($\mathcal{V}_N^*$) to explicitly model the perturbation, our method successfully decouples the noise from the identifier. The visualization shows that with this intervention, the model correctly learns the association between $\mathcal{V}^*$ and the personalized concept, generating a clean, high-fidelity image that accurately reflects the subject's identity from the same prompt. This directly illustrates how our method breaks the shortcut and restores the model's intended functionality. Additional visual results demonstrating concept decoupling are provided in Fig.~\ref{fig:causal-graph-noise}.

% From the above analysis, we can see that the injected perturbation, $\bdelta$, though imperceptible to human perception,
% We further distill such empirical
% However, another things to notice is that, the generated portraits actually are significantly low-quality than the perturbed images with the protective perturbation $\bdelta$.
% Injected protective perturbation, $\bdelta$, though imperceptible to human perception, indeed exploits the adversarial vulnerability of latent encoder, that the perturbed images actually larger diverged from its original clean person concept in latent space.
\blueUpdate{We refer to the path $\mathcal{V}^* \rightarrow \Delta$ as the identifier--noise shortcut.} To establish and reinforce this shortcut path, we find that effective perturbations must cause a \textit{latent-space image-prompt mismatch} (where latent space refers to the joint image--text representation space of the underlying CLIP encoders in diffusion models), where the perturbed images and their corresponding prompts are no longer semantically aligned. Learning from such mismatched pairs creates contradictions and encourages the model to encode chaotic perturbation patterns into the rarely used identifier token $\mathcal{V}^*$, instead of learning the clean identity behind $\bm{x}_0$.
% , which means the images and its corresponding prompts are not semantically aligned in the latent space after the perturbation.
% should cause is the \textit{latent-space image-prompt mismatch}, which means the images and its corresponding prompts are not semantically aligned in the latent space after the perturbation.
We infer this from two empirical observations: (i) random perturbation with the same strength does not affect the learning performance of the personalized diffusion model; (ii) the generated portraits using the perturbed diffusion model usually have lower quality and larger image distortion than the slightly perturbed input images. The first observation justifies that perturbations must cause a significant latent shift to affect the model's learning. The second suggests that the model learns abstract noise concepts rather than just pixel-level noise patterns. We further validate this through the following experiments of latent-mismatch visualization and concept interpretation.
%  First, using the paried CLIP image-text encoder in pretrained diffusion models, we conduct 2D visualization on the latent distribution of clean and perturbed images along with the person-related and noise-related concepts. Using three distinct visualization methods (TSNE, Truncated SVD and UMAP), we observe consistent results that the perturbed images and its corresponding prompts are not semantically aligned in the latent space.

\begin{figure}[!htb]
  \centering
  \includegraphics[width=\linewidth]{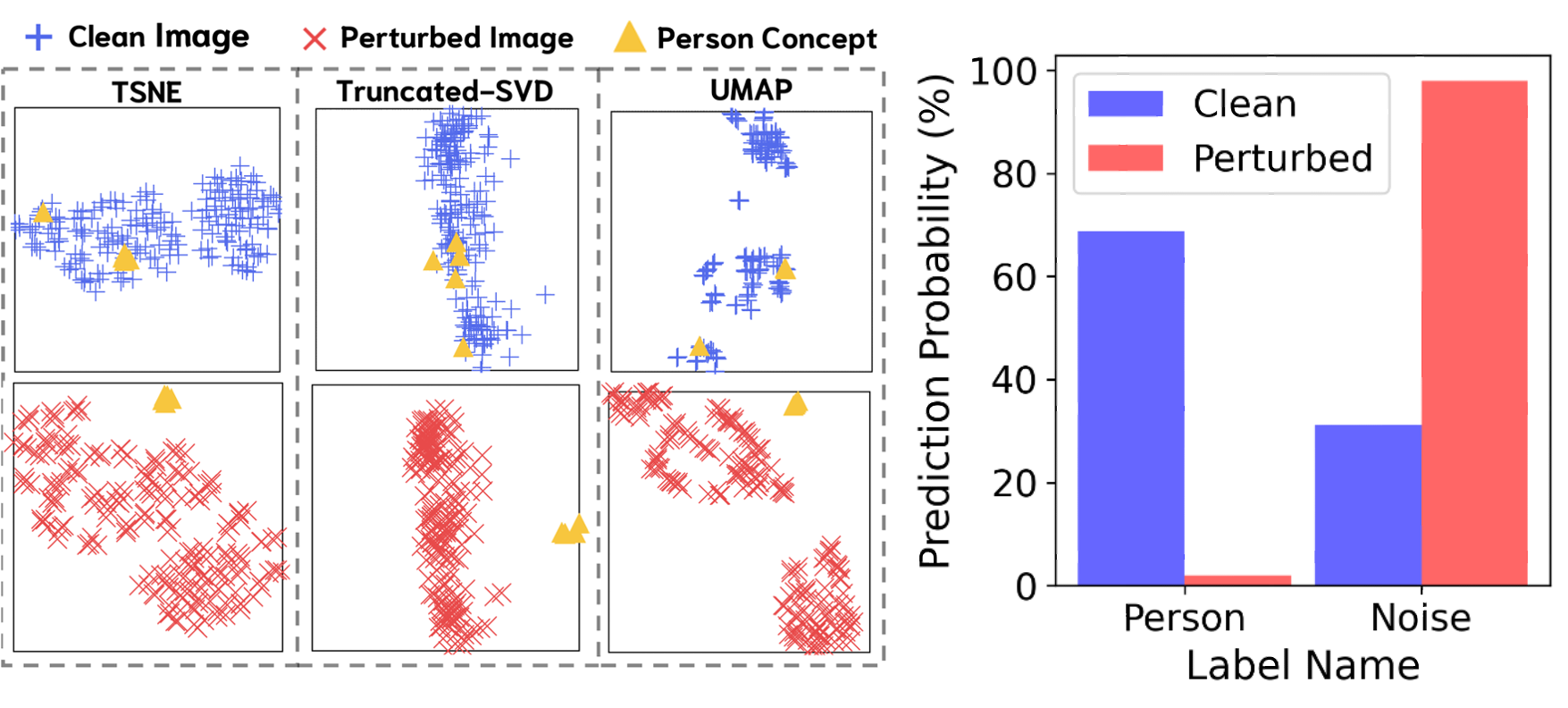}
  \caption{Latent 2D visualization and concept classification of images using CLIP encoders.}
  \label{fig:latent-vis}
\end{figure}

Specifically, we use the paired CLIP encoders to embed both the clean and perturbed images, as well as textual prompts describing the "person" concept (e.g., \textit{"a photo of a person's face"}). Then, we leverage three distinct 2D visualization techniques, including t-SNE~\citep{maaten2008visualizing}, Truncated-SVD~\citep{halko2011finding}, and UMAP~\citep{mcinnes2018umap} on image-prompt embedding pairs. The results in Fig. \ref{fig:latent-vis} suggest that \blueUpdate{protective} perturbation indeed significantly shifts the portrait latent from its original region of the ``person'' concept. Moreover, we precisely split the latent space into two regions with a zero-shot CLIP-based classifier, where we find that the perturbed images have a higher probability of being classified into the ``noise'' region instead of the ``person'' region in latent space. \blueUpdate{We provide more analysis on the learned concepts with the help of the causal graph in Fig.~\ref{fig:causal-graph}.}
% We hypothesize that learning on such perturbed images will create contradiction and force the models to dump that chaotic perturbation pattern into the rarely-appeared identifier token $\mathcal{V}^*$ instead of learning clean identity behind $\bm{x}_0$.
% As a result, linking the personal concept texutal prompt to the noise images will create contradiction and force the models to dump that chaotic perturbation pattern into the rarely-appeared identifier token $\mathcal{V}^*$ instead of learning clean identity behind $\bm{x}_0$.
% These findings indicate adversarial perturbation indeed leads to latent mismatch, which is also aligned with the previous studies that reveal the fundamental vulnerability of underlying latent encoders like VAE and CLIP encoders \citep{akrami2020robust, schlarmann2024robust, xue2023toward}.

These findings indicate \blueUpdate{protective} perturbation indeed leads to latent mismatch.
% , which is also aligned with the previous studies that reveal the fundamental vulnerability of underlying latent encoders like VAE and CLIP encoders \citep{akrami2020robust, schlarmann2024robust, xue2023toward}.
% Due to the natural constrain that the model need to neither link the unique identifier $\mathcal{V}^*$ to the noise, nor to the person identity concept $X_0$, this is when the shortcut learning happens. That is, the model is forced to learn the spurious correlations between $\mathcal{V}^*$ and $\delta(X_0)$ since it minimizes training loss in a more effective way for deep neural networks.
% Since the target $X_0$ is now greatly shifted semantically in latent,
% \blueUpdate{And this latent mismatch creates an opportunity for shortcut learning \citep{geirhos2020shortcut}, that is, the model prioritize the more available and easy-to-learn patterns for minimizing the training loss. In our case, the PDMs need to either link the unique identifier $\mathcal{V}^*$ to the noise $\Delta$, or to the person identity concept $X_0$. And from the perspective of more effective loss minimizing, the PDM ends up linking to the more high-frequency noise patterns, instead of the more desired person identity concept $X_0$.}
\blueUpdate{This latent mismatch creates an opportunity for shortcut learning~\citep{geirhos2020shortcut,hermann2023foundations}, where models optimize for easily accessible features rather than robust predictive patterns. In our case, PDMs face a binary choice: linking the unique identifier $\mathcal{V}^*$ either to the noise $\Delta$ or to the person identity concept $X_0$. From the perspective of loss minimization efficiency, PDMs naturally gravitate toward learning the high-frequency noise patterns rather than the more complex and desired person identity concept $X_0$, as this provides a computationally easier path to reduce training loss.}

\subsection{
  Training Clean PDMs on Perturbed Data with Systematic \blueUpdate{Red-Teaming}
  % Via Causal Intervention
% Systematic Defending with Input Purification, Decouple Learning, and Guided Sampling
}
To address this shortcut learning issue, we propose a systematic \blueUpdate{red-teaming} framework inspired by causal intervention \citep{geirhos2020shortcut}, which is a widely used technique to mitigate shortcut learning in traditional machine learning tasks.
% strategies used to mitigate shortcut learning in traditional machine learning tasks. In traditional machine learning, shortcut learning is often addressed by modifying the training data or the learning process to prevent models from exploiting spurious correlations \citep{geirhos2020shortcut}. Techniques such as data augmentation, adversarial training, and causal interventions encourage models to focus on the true underlying concepts rather than relying on shortcuts. For example, data augmentation introduces variability that forces the model to learn more robust features, while adversarial training exposes the model to perturbations during training to improve its resilience \citep{goodfellow2014explaining}.
Causal intervention \citep{kaddour2022causal} usually involves data augmentation or modifying the training process to disrupt spurious correlations.
% In our PDMs learning case, we propose the following strategies for mitigating shortcut learning: (i) \textit{Removing Noise Variables:} we employ off-the-shelf image restoration techniques to eliminate the direct influence of adversarial noise $\bdelta$. This step realigns the images with their true semantic representations in the latent space, disrupting the spurious correlations between noise patterns and unique identifiers; (ii) \textit{Weakening Spurious Paths and Strengthening Causal Paths:} we introduce \textit{Contrastive Decoupling Learning} to encourage the model to disentangle personalized concepts from noise patterns. By incorporating noise tokens into the prompts and designing contrastive learning objectives, we weaken the spurious associations that lead to poor generalization. Moreover, we enhance the learning of true causal relationships by leveraging clean prior data for emphasizing correct associations and conduct quality-enhanced sampling. We describe them in detail below and summarize the overall framework in Algorithm \ref{algo:alg_concise}.
To mitigate shortcut learning in PDMs, we propose two key strategies:
\begin{itemize}
    \item \textbf{Removing Noise Variables:} We use image restoration to eliminate adversarial noise, realigning images with their true semantic representations.
    \item \textbf{Weakening Spurious and Strengthening Causal Paths:} We introduce Contrastive Decoupling Learning, which uses noise tokens and clean prior data to disentangle personalized concepts from noise patterns.
\end{itemize}
We detail these approaches below and summarize our framework in Algorithm \ref{algo:alg_concise}. The causal interpretation is detailed in Section~\ref{sec:causal_intervention}.

% By systematically addressing the components contributing to shortcut learning, our approach ensures that the PDM learns robust and generalizable representations, even when trained on perturbed datasets. This methodology not only mitigates the effects of adversarial perturbations but also contributes to a deeper understanding of the causal mechanisms underlying PDM training.

% \subsection{Detailed Defense Strategies}

% In this section, we describe each of the proposed modules in detail and summarize the overall framework in Algorithm \ref{algo:alg_concise}.

\header{Image Purification via Image Restoration.} An intuitive and effective approach to removing the direct influence of adversarial noise is to purify the input images using image restoration techniques. We view the perturbed images as degraded images in the image restoration domain \citep{wang2021deep} and leverage off-the-shelf image restoration models to convert low-quality, noisy images into high-quality, purified ones. Specifically, we use a face-oriented model named CodeFormer \citep{zhou2022codeformer}, which is trained on facial data to restore images based on latent code discretization. To further enhance the purification of non-face regions, we employ an additional diffusion-based super-resolution (SR) model. Compared to previous state-of-the-art optimization-based purification methods \citep{cao2024impress} and diffusion-based purification methods \citep{zhao2024can}, this simple yet effective pipeline yields faithful purified images with better efficiency since it only requires a single inference pass. We term this module \textit{CodeSR} as it combines CodeFormer and SR in sequence.

\header{Contrastive Decoupling Learning (CDL).} To further mitigate shortcut learning, we introduce Contrastive Decoupling Learning, which aims to disentangle the learning of desired personalized concepts from undesired noise patterns. We achieve this by augmenting the prompts with additional tokens related to the noise pattern, denoted as $\mathcal{V}^*_N$, such as \textit{``t@j noisy pattern''}. Ideally, these newly added tokens absorb all the noise components in the image, leaving the clean, personalized concept associated with the personalized identifier $\mathcal{V}^*$. During training, we insert $\mathcal{V}^*_N$ into the prompt of instance data with the suffix \textit{``with t@j noisy pattern''}, and include the ``inverse'' of $\mathcal{V}^*_N$ in the prompt of class-prior data with the suffix \textit{``without t@j noisy pattern''}. This contrastive prompt design encourages the model to distinguish between the instance concept and noise patterns, thus weakening spurious correlations. During inference, we add the suffix \textit{``without t@j noisy pattern''} to the prompt input to guide the model in disregarding the learned patterns associated with $\mathcal{V}^*_N$, thereby generating images that focus on the personalized concept. Furthermore, we use classifier-free guidance~\citep{ho2022classifier} with a negative prompt ($c_{\text{neg}} = \textit{``noisy, abstract, pattern, low quality''}$) to guide the model toward generating higher-quality images. Specifically, given timestamp $t$, we perform sampling using the linear combination of the good-quality and bad-quality conditional noise estimates with guidance weight $w^{\text{neg}}=7.5$:
% \vspace{-2pt}
\begin{equation}
\label{eq:qfs}
\tilde{\bm{\epsilon}}_\theta\left(\mathbf{z}_t, \mathbf{c}\right) = (1 + w^{\text{neg}}) \bm{\epsilon}_\theta\left(\mathbf{z}_t, \tau_{\theta}(c^{[\mathcal{V}^*, \bar{\mathcal{V}}_N^*]})\right) - w^{\text{neg}} \bm{\epsilon}_\theta\left(\mathbf{z}_t, \tau_{\theta}(c_{\text{neg}})\right)
\end{equation}
where $\bar{\mathcal{V}}_N^*$ denotes the negation of the noise token (i.e., using ``without $\mathcal{V}_N^*$'' in the prompt), and $\tau_{\theta}(\cdot)$ is the text encoder.

% \header{Strengthening Causal Paths.} To reinforce the learning of true causal relationships, we leverage clean prior data $\bar{\xb}_0$ and emphasize correct associations between images and text prompts. By incorporating clean data during training and carefully designing prompts, we ensure that the model focuses on learning the desired personalized concepts without being influenced by spurious noise patterns.

% Additionally, during inference, we employ quality-enhanced sampling strategies, such as the negative prompting described above, to generate high-quality images that accurately reflect the learned concepts. This approach helps the model to generalize better and produce outputs that are consistent with the true underlying data distribution.

\begin{algorithm}[t]
  \caption{Training Clean Personalized LDMs on Perturbed Data with Systematic \blueUpdate{Red-Teaming}}
  \label{algo:alg_concise}
  \begin{algorithmic}[1]
  \Require Corrupted training set $X^{\prime}_0$, pre-trained LDM $\theta_0$, CodeFormer $\phi=\{\mathcal{E}_\phi, \mathcal{D}_\phi, \mathcal{T}_\phi, \mathcal{C}\}$, SR model $\psi$, prior data $\bar{X}_0$, noise token $\mathcal{V}_N^*$, personalized identifier $\mathcal{V}^*$, instance prompt $c^{\mathcal{V}^*}$, class prompt $c$, number of generations $N_{\text{gen}}$

  \Ensure Personalized diffusion model with clean-level generation performance $\theta_T$

      \State \textbf{Step 1: Input Purification with CodeFormer and Super-resolution Model}

      \State \textit{CodeFormer}: Predict code $\tilde{Z}_c = \mathcal{T}_{\phi}( \mathcal{E}_\phi(X^{\prime}_0), \mathcal{C})$; obtain high-quality restoration $\tilde{X}_0 = \mathcal{D}_\phi(\tilde{Z}_c)$

      \State \textit{Super-resolution}: Resize $\tilde{X}_0$ to $128 \times 128$; apply SR model $\psi$ to obtain $\tilde{X}_0^{\text{purified}}$ at $512 \times 512$

      \State \textbf{Step 2: Contrastive Decoupling Learning}
      \For{$i = 1$ \textbf{to} $T$ training steps}

          \State Sample instance data $x_i$ from $\tilde{X}_0^{\text{purified}}$, and class-prior data $\bar{x}_0$ from $\bar{X}_0$

          \State Craft decoupled instance prompt $c^{\mathcal{V}^*}_{\text{dec}} = \texttt{concat}(c^{\mathcal{V}^*}, \mathcal{V}_N^*)$ and class-prior prompt $c_{\text{dec}} = \texttt{concat}(c, \text{``without''}, \mathcal{V}_N^*)$

          \State Optimize the LDM $\theta_i$ with standard DreamBooth loss $\mathcal{L}_{\text{db}}$ \Comment{Following Eq.~\ref{eq:DB_loss}}
          \State $\mathcal{L}_{\text{db}}(x_i, c^{\mathcal{V}^*}_{\text{dec}}, \bar{x}_0, c_{\text{dec}}; \theta_i ) = \mathcal{L}_{\mathrm{denoise}}\left( x_i, c^{\mathcal{V}^*}_{\text{dec}} \right) + \lambda \mathcal{L}_{\mathrm{denoise}}\left( \bar{x}_0, c_{\text{dec}} \right)$

          \State Update LDM $\theta_i$ with $\nabla_{\theta_i} \mathcal{L}_{\text{db}}$ using AdamW optimizer on U-Net Denoiser and Text Encoder

      \EndFor

      \State \textbf{Inference:} Perform decoupled sampling $\{X_{\text{gen}}^{j}\}_{j=1}^{N_{\text{gen}}}$ with the trained PDM \Comment{Following Eq.~\ref{eq:qfs}}

  \end{algorithmic}
\end{algorithm}

% Math packages and definitions
\section{Experiments}
\label{sec:exp}
% In this section, we first perform an evaluation of the effectiveness, efficiency, and faithfulness of the proposed purification techniques. Then, we investigate the resilience of our framework under adaptive perturbation against the purification models. Lastly, we perform systematic ablation studies and analysis to provide an in-depth exploration of each module in our framework.

\subsection{Experimental Setup}

\label{sec:setup}

\header{Datasets and Metrics.} We mainly conduct quantitative experiments on {VGGFace2} \citep{cao2018vggface2} following \citep{van2023anti,Liu_2024_CVPR}. We select four identities, randomly pick eight images from each individual, and split them into two subsets for image protection and reference.
Additionally, we visually demonstrate the purification ability on samples from an artwork dataset, WikiArt \citep{saleh2015large}, and the CelebA \citep{liu2015deep}.
For the metric, we evaluate the generated images in terms of their \textit{semantic-related quality} and \textit{graphical aesthetic quality}. For the semantic-related score, we compute the cosine similarity between the embedding of generated images and reference images, which we term the Identity Matching Similarity (IMS) score. IMS ranges from $-1$ to $1$, where higher values indicate greater semantic alignment with the reference identity.
We report a weighted average IMS score by employing two face embedding extractors, including \textit{antelopev2} model from InsightFace library \citep{Deng2020CVPR} following IP-adapter \citep{ye2023ip-adapter} and \textit{VGG-Net} \citep{simonyan2014very} from DeepFace library \citep{serengil2021lightface} following \citep{van2023anti}.
The IMS score is computed via a weighted sum: IMS$=\lambda \text{IMS}_{\text{IP}}+(1-\lambda)\text{IMS}_{\text{VGG}}$, where $\lambda$ is set as 0.7.
For the graphical quality $Q$, we report the average of two metrics: i) \textit{LIQE} \citep{zhang2023blind} (with re-normalization to $[-1,+1]$); ii) \textit{CLIP-IQAC} following \citep{Liu_2024_CVPR}, which is based on CLIP-IQA \citep{wang2022exploring} with class label.

\header{Purification Baselines and Perturbation Methods. }
For \textit{purification baselines}, we consider both model-free and diffusion-based approaches. Model-free methods operate with image processing algorithms and, despite their simplicity, can achieve non-trivial defense performance against adversarial and availability attacks~\citep{liang2023adversarial,Liu_2024_CVPR}. Diffusion-based methods leverage powerful diffusion probabilistic models that have strong ability to model real-world data distribution and show potential as zero-shot purifiers~\citep{cao2024impress,zhao2024can}.
\begin{itemize}[noitemsep,leftmargin=*]
    \item \textbf{Model-free methods:} We include Gaussian Filtering (kernel size 5), which smooths high-frequency adversarial perturbations; Total Variation Minimization (TVM)~\citep{wang2020adversarial}, which reconstructs images based on the observation that benign images have low total variation; and JPEG Compression (quality 75), which reduces file size and removes perturbation artifacts.
    \item \textbf{Diffusion-based methods:} We include (Pixel)DiffPure~\citep{nie2022diffusion}, which diffuses adversarial examples with noise and recovers clean images through the reverse process; a latent-space variant (LatentDiffPure) using Stable Diffusion; DDSPure~\citep{carlini2022certified,hu2023improving}, which finds optimal timestamps for perturbation removal via SDEdit; GrIDPure~\citep{zhao2024can}, which conducts iterative purification on image grids; and IMPRESS~\citep{cao2024impress}, which ensures latent consistency with visual similarity constraints.
\end{itemize}
For \textit{protective perturbation}, we consider seven state-of-the-art approaches that either craft noise against fixed LDMs or jointly learn the noise generator and perturbation for better protection capacity~\citep{van2023anti,Liu_2024_CVPR}. Bi-level optimization methods include \textit{FSMG}, which uses a fully-trained surrogate model; \textit{ASPL}, which alternates surrogate training with perturbation learning; \textit{EASPL}~\citep{van2023anti}, which uses an ensemble of surrogates for better transferability; and \textit{MetaCloak}~\citep{Liu_2024_CVPR}, which employs meta-learning with transformation sampling for robust perturbations. Fixed-model methods include \textit{AdvDM}~\citep{liang2023adversarial}, which maximizes denoising error in an untargeted way; \textit{PhotoGuard}~\citep{salman2023raising}, which uses target-adversarial perturbations for the encoder or full diffusion model; and \textit{Glaze}~\citep{shan2023glaze}, which transfers artwork to a target style to prevent style learning. We set ASPL as the default perturbation with $\ell_\infty$ radius of 11/255 and six-step PGD with step size 1/255 following \citep{van2023anti}.

\begin{table*}[t]
  \centering
  \caption{
  Comprehensive evaluation of different purification methods under various protective perturbations. Higher values are better for both Identity Matching Similarity (IMS) and Quality (Q) metrics. Best results are in \textbf{bold}, and second-best results are shaded. $*$ denotes statistical significance (p $\leq$ 0.01) in Wilcoxon signed-rank test.
  }
  \label{tab:main_res}
\resizebox{\linewidth}{!}{
  \begin{tabular}{@{\extracolsep{\fill}}ccccccccccccccccc}
    \toprule
    \toprule
    \multirow{2}{*}{\textbf{Methods} } & \multicolumn{2}{c}{\textbf{Clean}} & \multicolumn{2}{c}{\textbf{FSMG}} & \multicolumn{2}{c}{\textbf{ASPL}} & \multicolumn{2}{c}{\textbf{EASPL}} & \multicolumn{2}{c}{\textbf{MetaCloak}} & \multicolumn{2}{c}{\textbf{AdvDM}} & \multicolumn{2}{c}{\textbf{PhotoGuard}} & \multicolumn{2}{c}{\textbf{Glaze}} \\
    \cmidrule(lr){2-3}\cmidrule(lr){4-5}\cmidrule(lr){6-7}\cmidrule(lr){8-9}\cmidrule(lr){10-11}\cmidrule(lr){12-13}\cmidrule(lr){14-15} \cmidrule(lr){16-17}
    & \textbf{IMS} $\uparrow$ & \textbf{Q} $\uparrow$ & \textbf{IMS} $\uparrow$ & \textbf{Q} $\uparrow$ & \textbf{IMS} $\uparrow$ & \textbf{Q} $\uparrow$ & \textbf{IMS} $\uparrow$ & \textbf{Q} $\uparrow$ & \textbf{IMS} $\uparrow$ & \textbf{Q} $\uparrow$ & \textbf{IMS} $\uparrow$ & \textbf{Q} $\uparrow$ & \textbf{IMS} $\uparrow$ & \textbf{Q} $\uparrow$ & \textbf{IMS} $\uparrow$ & \textbf{Q} $\uparrow$ \\
    \midrule
    \textbf{Clean}
     & -0.13 & 0.15  & -0.13 & 0.15& -0.13 & 0.15& -0.13 & 0.15& -0.13 & 0.15& -0.13 & 0.15& -0.13 & 0.15& -0.13 & 0.15 \\
    \textbf{Perturbed}
    & - & -  &-0.43 & -0.54& -0.67 & -0.52& -0.62 & -0.50& -0.35 & -0.53& -0.27 & -0.36& -0.18 & -0.24& -0.28 & -0.28 \\
    \midrule
    \textbf{Gaussian F.}   & -0.23 & -0.52 &-0.19 & -0.55& -0.20 & -0.57& -0.17 & -0.58& \secondrunner{-0.07} & -0.63& -0.11 & -0.57& -0.23 & -0.53& -0.18 & -0.54 \\
    \textbf{JPEG}  & -0.27 & -0.13 & -0.15 & -0.41& -0.21 & -0.52& -0.27 & -0.50& -0.34 & -0.38& -0.15 & \secondrunner{-0.02}& -0.13 & \secondrunner{0.07}& -0.19 & \secondrunner{-0.03} \\
    \textbf{TVM}  & -0.15 & -0.64 & -0.12 & -0.65& -0.16 & -0.66& -0.10 & -0.67& -0.11 & -0.69& -0.12 & -0.65& -0.15 & -0.64& -0.11 & -0.66 \\
    \textbf{PixelDiffPure}  & -0.34 & -0.60 & -0.41 & -0.57& -0.43 & -0.54& -0.57 & -0.61& -0.28 & -0.58& -0.40 & -0.55& -0.25 & -0.55& -0.41 & -0.59 \\
    \textbf{L.DiffPure-$\emptyset$}  & -0.24 & 0.16 & \secondrunner{-0.07} & -0.47 & -0.36 & -0.59 & -0.22 & -0.49 & -0.52 & -0.43 & -0.55 & -0.24 & -0.12 & -0.40 & -0.38 & -0.42 \\
    \textbf{L.DiffPure}  & -0.28 & \secondrunner{0.21} & -0.25 & -0.45 & -0.31 & -0.61 & -0.30 & -0.46 & -0.31 & -0.51 & -0.57 & -0.30 & -0.25 & -0.47 & -0.41 & -0.47 \\
    \textbf{DDSPure}  & -0.25 & -0.38 & -0.15 & -0.34 & \secondrunner{-0.05} & -0.38 & \secondrunner{-0.08} & -0.39 & -0.16 & -0.49 & -0.19 & -0.43 & -0.12 & -0.37 & -0.22 & -0.41 \\
    \textbf{GrIDPure}  & -0.46 & -0.17 & -0.10 & \secondrunner{-0.20} & -0.21 & \secondrunner{-0.16} & -0.13 & \secondrunner{-0.25} & -0.23 & \secondrunner{-0.25} & \secondrunner{-0.09} & -0.18 & \secondrunner{-0.03} & -0.22 & -0.24 & -0.13 \\
    \textbf{IMPRESS}  & \secondrunner{-0.02} & -0.18 & -0.15 & -0.53 & -0.16 & -0.49 & -0.29 & -0.64 & -0.34 & -0.29 & -0.34 & -0.34 & -0.16 & -0.21 & \secondrunner{-0.10} & -0.43 \\
    \midrule
    \textbf{Ours}  & \textbf{0.14$^*$} & \textbf{0.54$^*$} & \textbf{0.23$^*$} & \textbf{0.65$^*$} & \textbf{0.09} & \textbf{0.62$^*$} & \textbf{0.09$^*$} & \textbf{0.63$^*$} & \textbf{0.38$^*$} & \textbf{0.58$^*$} & \textbf{0.29$^*$} & \textbf{0.67$^*$} & \textbf{0.24$^*$} & \textbf{0.63$^*$} & \textbf{0.31$^*$} & \textbf{0.66$^*$} \\
    \bottomrule
    \bottomrule
    \end{tabular}
}
\end{table*}

\subsection{Effectiveness, Efficiency, and Faithfulness}
\header{Effectiveness Comparison.}
We present the effectiveness of different purification across seven perturbation methods in Tab. \ref{tab:main_res}. From the table, we can see that compared to the clean case, training on perturbed data causes severe model degradation in both identity similarity and image quality. Across all perturbations, ASPL causes the most severe degradation when no purification is used. In contrast, MetaCloak is more robust against rule-based purification. Compared to rule-based purification, diffusion-based approaches achieve better performance in improving both identity similarity and image quality in most settings. Among them, GrIDPure yields relatively better purification performance since it considers the structure consistency, which suppresses the generative nature during the purification. However, there are still gaps in the IMS score compared to the clean case, and most of the quality scores after conducting GrIDPure purification are still negative. Compared to these baselines, our method closes the gap by further improving the IMS and quality scores, which are even higher than the clean training case in all the settings. The reasons are twofold: first, we use image-restoration-based approaches, which preserve the image structure well; furthermore, our CDL module contributes significantly to quality improvement.

% Full comparison results with standard deviations are available in the extended version.

\begin{figure*}[t]
    \setlength{\tabcolsep}{1pt}
    \renewcommand{\arraystretch}{0.5}
    \begin{center}
     \resizebox{\linewidth}{!}{
    \begin{tabular}{ccccccc}
        \toprule
      \textbf{Perturbed}  & \textbf{LatentDiffPure-$\emptyset$} & \textbf{LatentDiffPure} &\textbf{DDSPure} & \textbf{GrIDPure} & \textbf{IMPRESS} & \textbf{Ours} \\
        \midrule
        \includegraphics[width=0.22\linewidth]{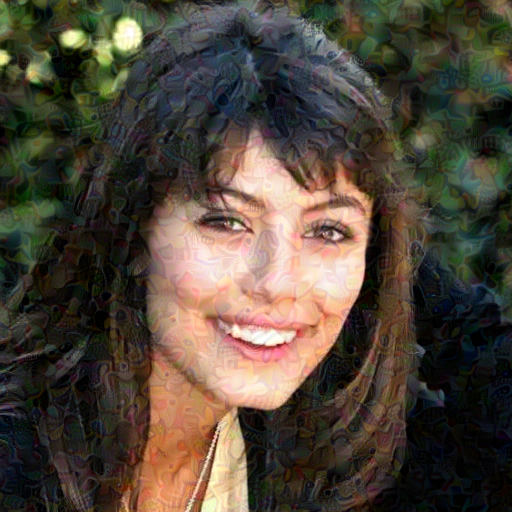} &
        \includegraphics[width=0.22\linewidth]{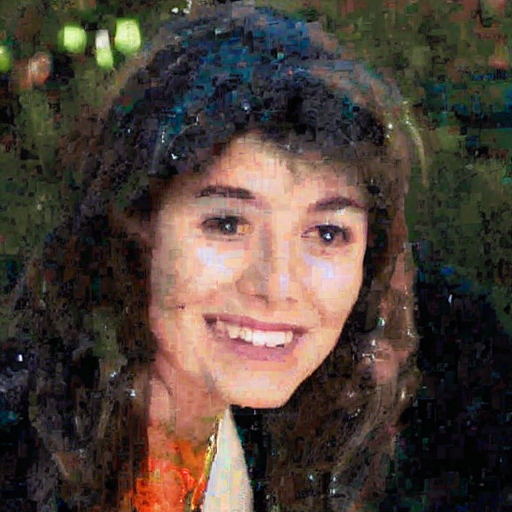} &
        \includegraphics[width=0.22\linewidth]{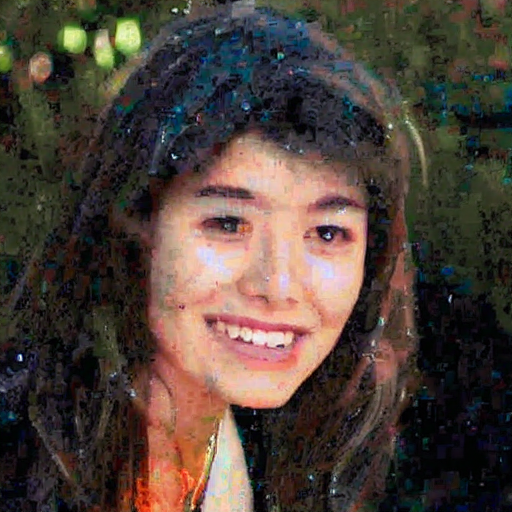} &
        \includegraphics[width=0.22\linewidth]{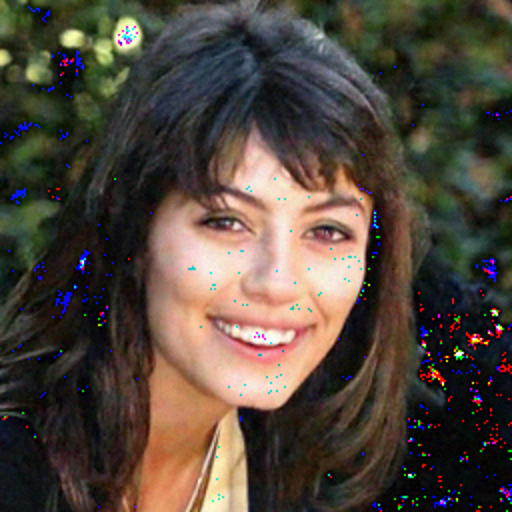} &
        \includegraphics[width=0.22\linewidth]{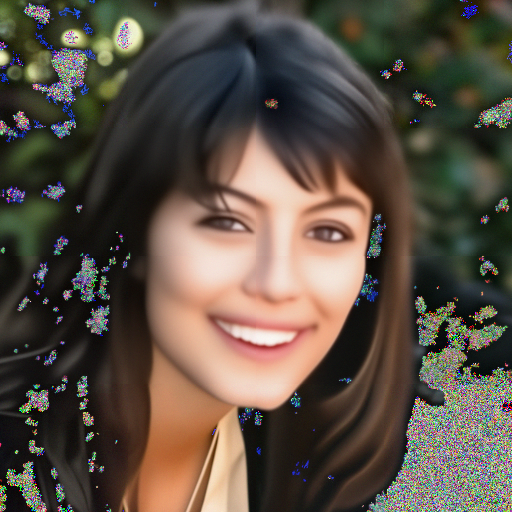} &
        \includegraphics[width=0.22\linewidth]{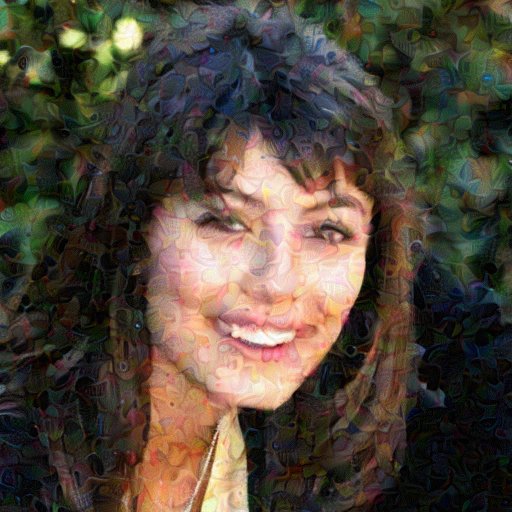} &
        \includegraphics[width=0.22\linewidth]{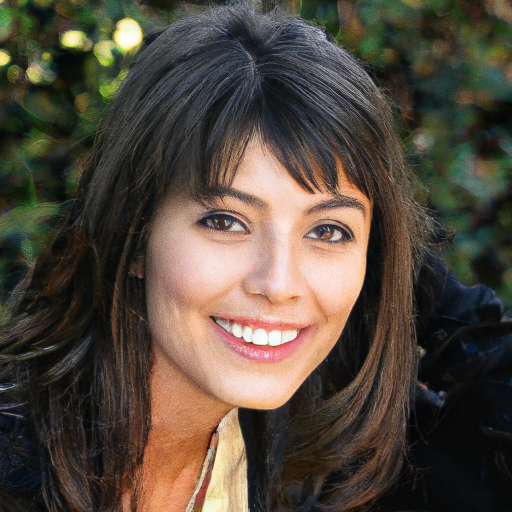}  \\
        \includegraphics[width=0.22\linewidth]{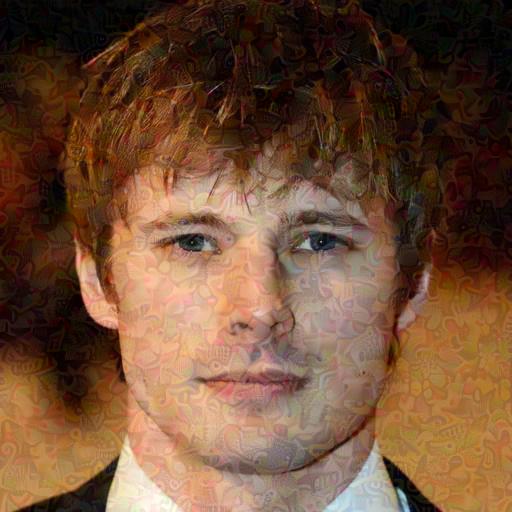} &
        \includegraphics[width=0.22\linewidth]{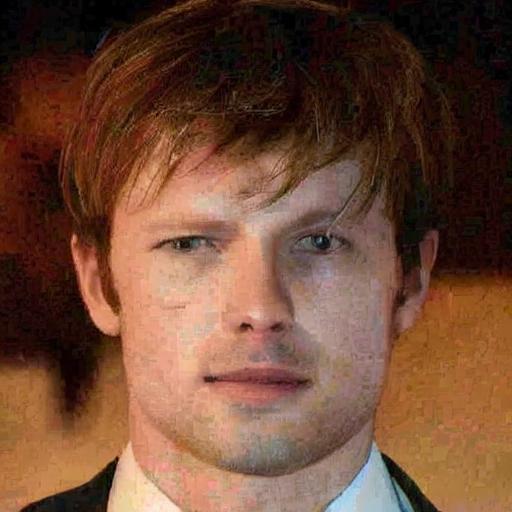} &
        \includegraphics[width=0.22\linewidth]{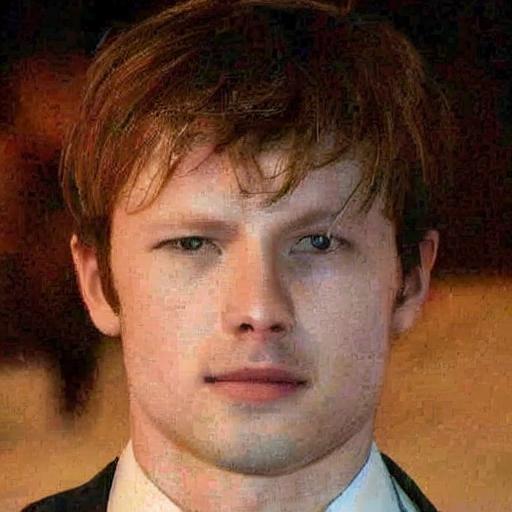} &
        \includegraphics[width=0.22\linewidth]{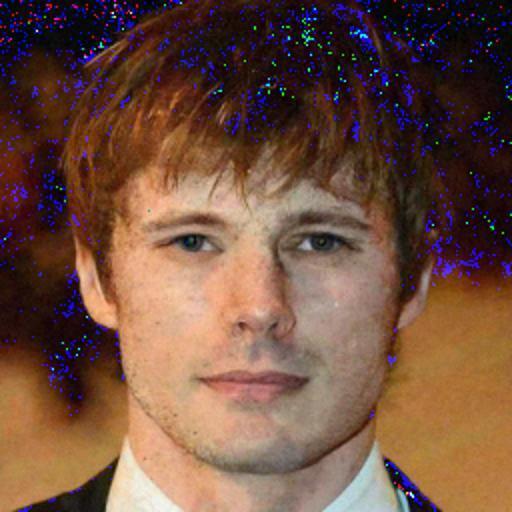} &
        \includegraphics[width=0.22\linewidth]{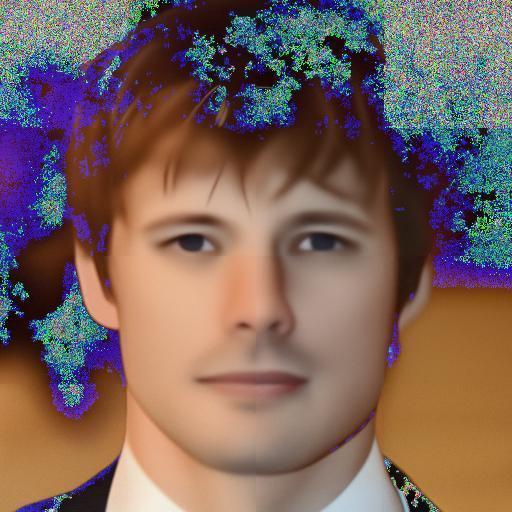} &
        \includegraphics[width=0.22\linewidth]{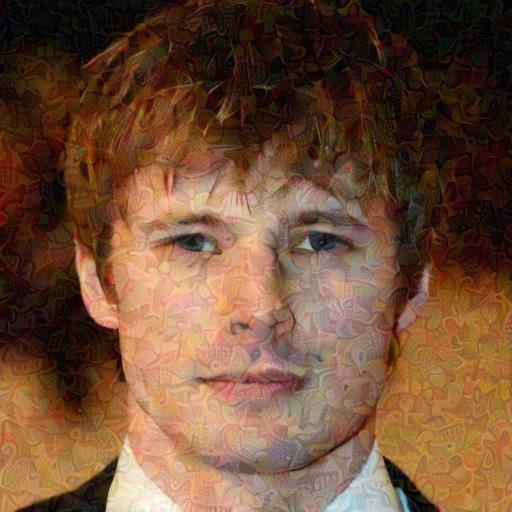} &
        \includegraphics[width=0.22\linewidth]{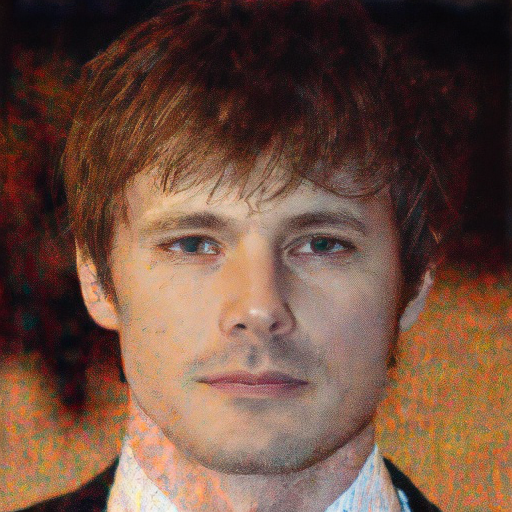}  \\
        \includegraphics[width=0.22\linewidth]{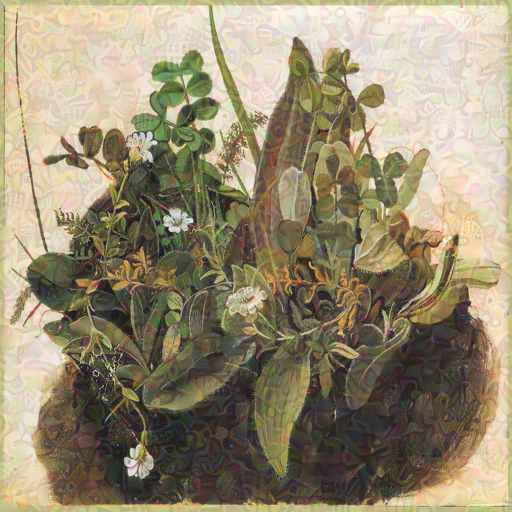} &
        \includegraphics[width=0.22\linewidth]{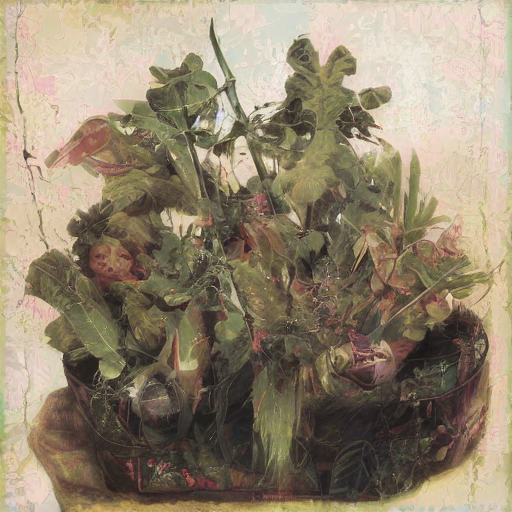} &
        \includegraphics[width=0.22\linewidth]{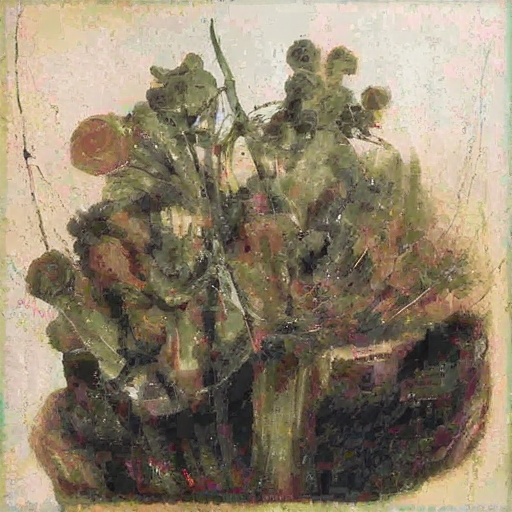} &
        \includegraphics[width=0.22\linewidth]{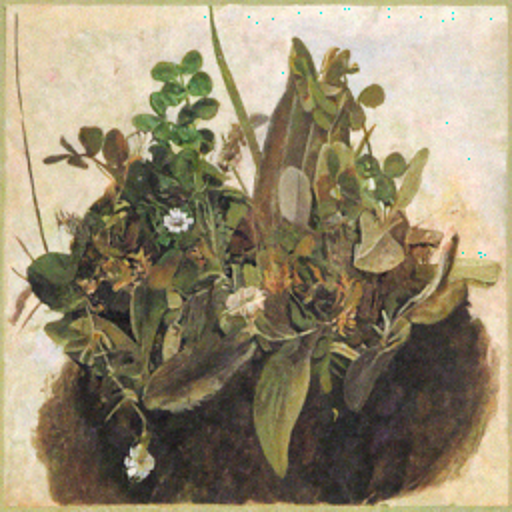} &
        \includegraphics[width=0.22\linewidth]{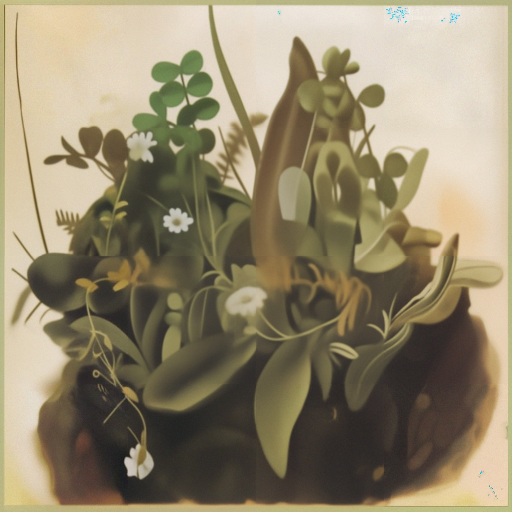} &
        \includegraphics[width=0.22\linewidth]{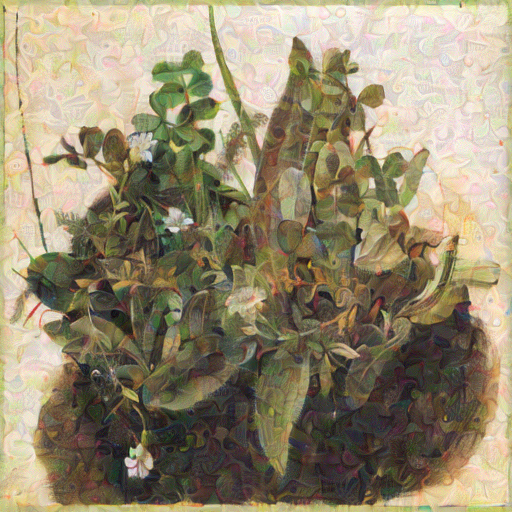} &
        \includegraphics[width=0.22\linewidth]{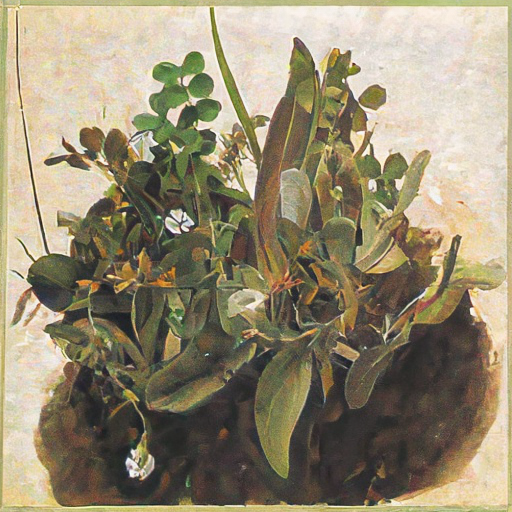}  \\
    \end{tabular}}
    \caption{Visualization of purified images that were originally protected by MetaCloak. Our method shows high faithfulness and high quality, while others fail to effectively purify the perturbation.}
    \label{vis}
    \end{center}
\end{figure*}

\header{Efficiency and Faithfulness of Purification.}
We present the evaluation of time cost and purification faithfulness compared to all other diffusion-based purification approaches in Tab. \ref{tab:faith-time}. The time cost is measured in seconds per sample with consideration of model loading.
Compared to other methods,
our purification has the lowest time cost and is 10$\times$ faster than the previous SoTA method, IMPRESS.
The reason behind this is that we leverage the super-resolution module, which empowers the usage of skip-step sampling to boost the generation time.
Moreover, we test the purification faithfulness of each method in terms of LPIPS loss~\citep{lpips}, a common metric measuring the visual perception distance of two images. From Tab. \ref{tab:faith-time}, we can see that our method achieves the lowest LPIPS loss. To visually validate this, we additionally present the purified images in Fig. \ref{vis}. From the figure, we can see that other diffusion-based approaches have limitations in hallucinating the content, introducing severe artifacts, or not having enough purification strength. In particular, we observed that LatentDiffPure causes significant identity changes during purification, likely due to semantic distortion in the latent space. On the other hand, GrIDPure \citep{zhao2024can} brings some artifacts to the purified image, which indicates that the underlying unconditional diffusion model pre-trained on ImageNet might not be suitable for general domain purification. In comparison, our purification method significantly enhances faithfulness by leveraging off-the-shelf image restoration models. These models are designed to preserve the structural integrity of the input, resulting in output images that closely maintain the original composition while effectively removing perturbations. This approach ensures that the purified images retain the essential features and identity of the original subjects, while successfully mitigating unwanted artifacts or noise.

\begin{table}[t]
  \centering
  \caption{Comparison of faithfulness (LPIPS, lower is better) and computational efficiency (seconds per sample) across different diffusion-based purification methods. Best results are in \textbf{bold}.}
  \label{tab:faith-time}
\resizebox{.6\linewidth}{!}{
  \begin{tabular}{l|cc}
    \toprule
    \textbf{Methods} & \textbf{LPIPS}$\downarrow$ & \textbf{Time (s)}$\downarrow$ \\
    \midrule
    IMPRESS & 0.451 & 675.0 \\
    PixelDiffPure & 0.495 & 102.0 \\
    DDSPure & 0.384 & 122.5 \\
    GrIDPure & 0.429 & 92.8 \\
    LatentDiffPure & 0.453 & 63.8 \\
    LatentDiffPure-$\emptyset$ & 0.450 & 63.3 \\
    \midrule
    \textbf{Ours} & \textbf{0.271} & \textbf{51.0} \\
    \bottomrule
    \end{tabular}
}
\end{table}

\begin{table}[t]
  \centering
  \caption{
    Analysis of model variants' effectiveness against adaptive attacks (AA). Results show IMS and Quality (Q) scores before and after AA, with performance degradation ($\Delta$). Best results in each column are in \textbf{bold}.
    }
  \label{tab:setting2}
\resizebox{\linewidth}{!}{
  \begin{tabular}{l|c|ccc|cccc}
    \toprule
    \multirow{2}{*}{\textbf{Modules}} & \multirow{2}{*}{\textbf{CDL}} & \multicolumn{3}{c|}{\textbf{Before AA}} & \multicolumn{4}{c}{\textbf{After AA}} \\
    \cmidrule{3-9}
     & & \textbf{IMS} & \textbf{Q} & \textbf{Avg.} & \textbf{IMS} & \textbf{Q} & \textbf{Avg.} & \textbf{$\Delta$} \\
    \midrule
    \multirow{2}{*}{\textbf{CodeSR}} & \xmarkcolor & \textbf{0.18} & 0.41 & \textbf{0.29} & -0.06 & 0.03 & -0.01 & $\downarrow$0.30 \\
     & \cmarkcolor & 0.23 & -0.03 & 0.10 & -0.09 & \textbf{0.24} & \textbf{0.08} & $\downarrow$\textbf{0.02} \\
    \midrule
    \multirow{2}{*}{\textbf{Code}} & \xmarkcolor & 0.14 & \textbf{0.42} & 0.28 & -0.09 & -0.17 & -0.13 & $\downarrow$0.41 \\
     & \cmarkcolor & 0.05 & 0.11 & 0.08 & -0.16 & -0.47 & -0.32 & $\downarrow$0.40 \\
    \midrule
    \multirow{2}{*}{\textbf{SR}} & \xmarkcolor & 0.07 & -0.02 & 0.03 & \textbf{0.20} & -0.39 & -0.09 & $\downarrow$0.12 \\
     & \cmarkcolor & 0.07 & -0.02 & 0.03 & -0.03 & 0.08 & 0.02 & $\downarrow$\textbf{0.01} \\
    \bottomrule
    \end{tabular}
  }
\end{table}

\subsection{Resilience Against Adaptive Perturbations }
\label{app. resilience}

DNN-based purification is prone to further adaptive attacks due to the non-smoothness in terms of latent representation space \citep{guo2024smooth} and also the vulnerability by exploiting adversarial examples \citep{ilyas2019adversarial}. We tested our framework's resilience to adaptive attacks that have knowledge of our pipeline. To do this, we evaluated different variants of our approach before and after applying an adaptive perturbation specifically targeting the image purification component. The adversarial perturbation is crafted following AdvDM with consideration of the CFG \citep{ho2022classifier} sampling trajectory with a large perturbation budget of $r=16/255$. For the model variants, we consider the full variant with all three modules enabled, as well as the ablated versions with one of them turned off. From Tab. \ref{tab:setting2}, we can see that the full variant with CDL is more robust to the adaptive attack than other variants in terms of performance drop. Furthermore, we notice that the variant with both SR and CDL yields a slightly better average score than the CodeSR configuration after the attack. This indicates that the CodeFormer module might be more susceptible to the adaptive attack while the SR module is more robust. However, using SR with CDL solely in such cases gives sub-optimal purification results. Our CodeSR configuration with CDL gives a better expected overall performance under mixed perturbation scenarios with $P$(AA)=50\%.

\subsection{Ablation Study and Sensitivity Analysis}
\begin{figure}[h]
  \centering
  \includegraphics[width=.8\linewidth]{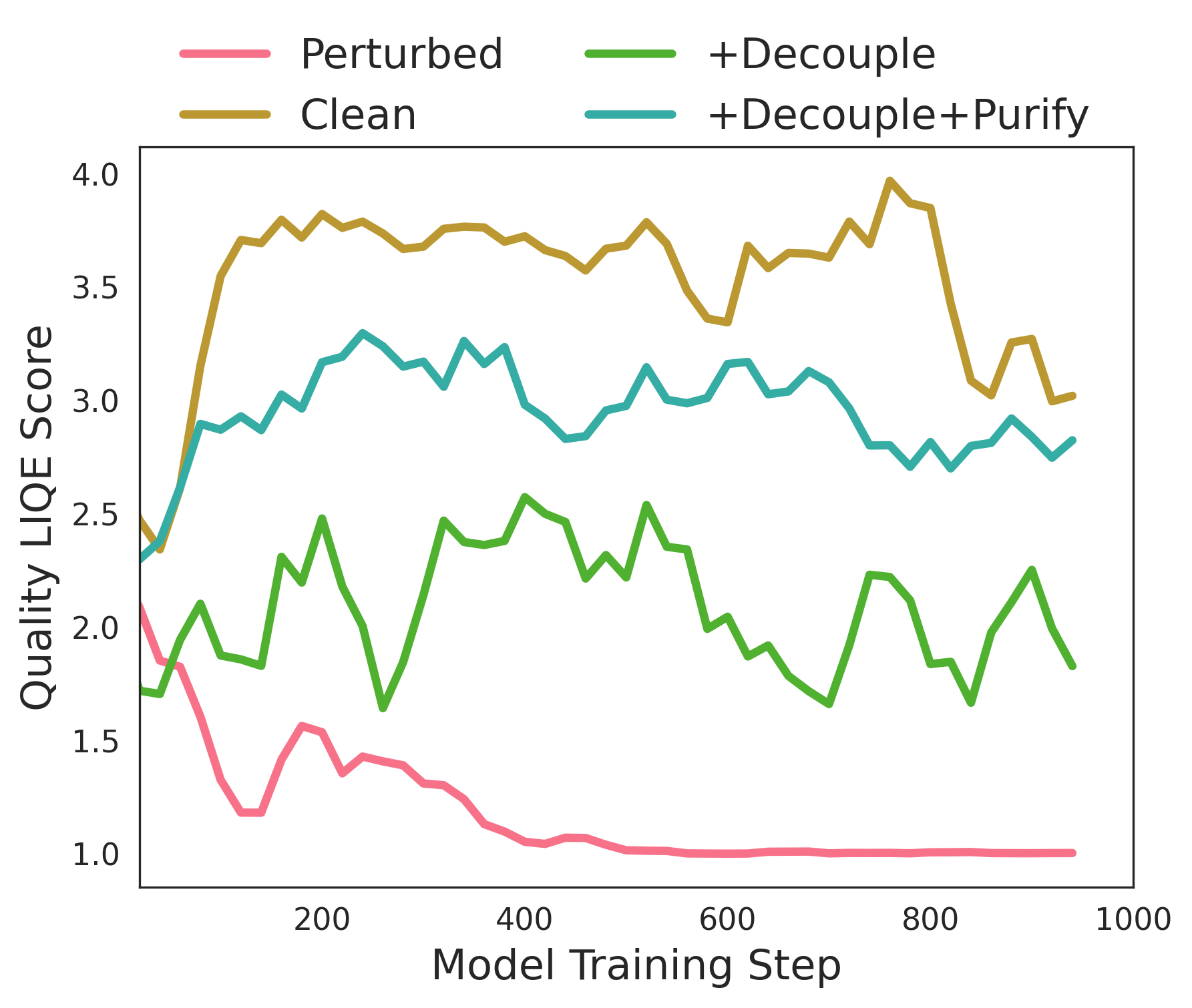}
  \caption{LIQE quality score curve of identifier $\mathcal{V}*$ during training. Our proposed decoupled learning (CDL) approach significantly enhances the quality compared to the case with perturbations. When combined with input purification (CodeSR + CDL), the model achieves quality performance comparable to clean-level training.}
  \label{fig:qualityloss}
\end{figure}

\header{Contribution of Individual Modules.} We present ablations on the three modules in our method in Tab.~\ref{tab:ablation}. From the table, our method works best under the full setting. When turning off any of the modules, the average performance degrades, with turning off CDL suffers the most. On the other hand, if we only turn on one of the modules, we find that CDL is still the most important one that retains higher generation performance. Furthermore, if we only do input purification without CDL, the generation quality is not as good as the full setting with CDL. This indicates that CDL is crucial for the performance of our method. Surprisingly, when only enabling the SR module, the IMS score is relatively good, but the visual quality is poor. While turning on the CodeFormer module alone, the boost is more on the quality score side. The settings that enable these two modules together yield a higher average score. This indicates that the SR and CodeFormer modules are complementary to each other.
Furthermore, for configurations that only allow two modules, we find that combining CodeFormer and CDL gives the best performance compared to the other two pairs. We also visualize the quality score curve of identifier $\mathcal{V}^*$ during training in Fig.~\ref{fig:qualityloss}, which shows consistent improvement.
Overall, the results suggest that all modules contribute to gains in both IMS and quality.

% === FIGURES MOVED FROM APPENDIX TO MAIN BODY (v1) ===

% Figure 6: Generation Visualization (moved from appendix/details.tex)
\begin{figure*}[t]
  \centering
  \includegraphics[width=0.98\linewidth,height=3.6cm]{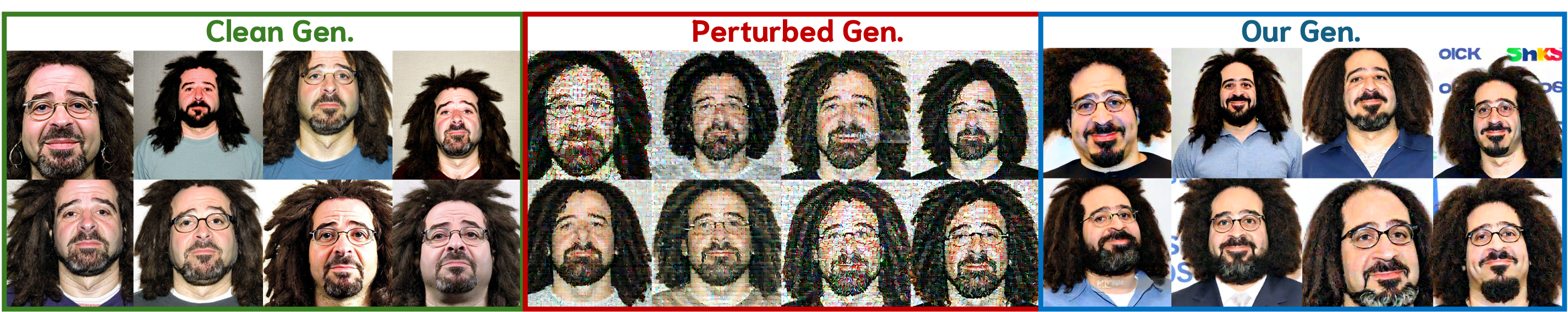}
  \caption{
  Generations from models trained on: (left) clean data, (middle) perturbed data without defense, and (right) purified data using our defense approach. The results demonstrate that our defense method significantly enhances generation quality, bringing it closer to clean data levels.
  }
  \label{fig:gen-vis}
\end{figure*}

% Figure 7: Causal Intervention Visualization (moved from appendix/analysis.tex)
\begin{figure}[t]
    \centering
    \includegraphics[width=\linewidth]{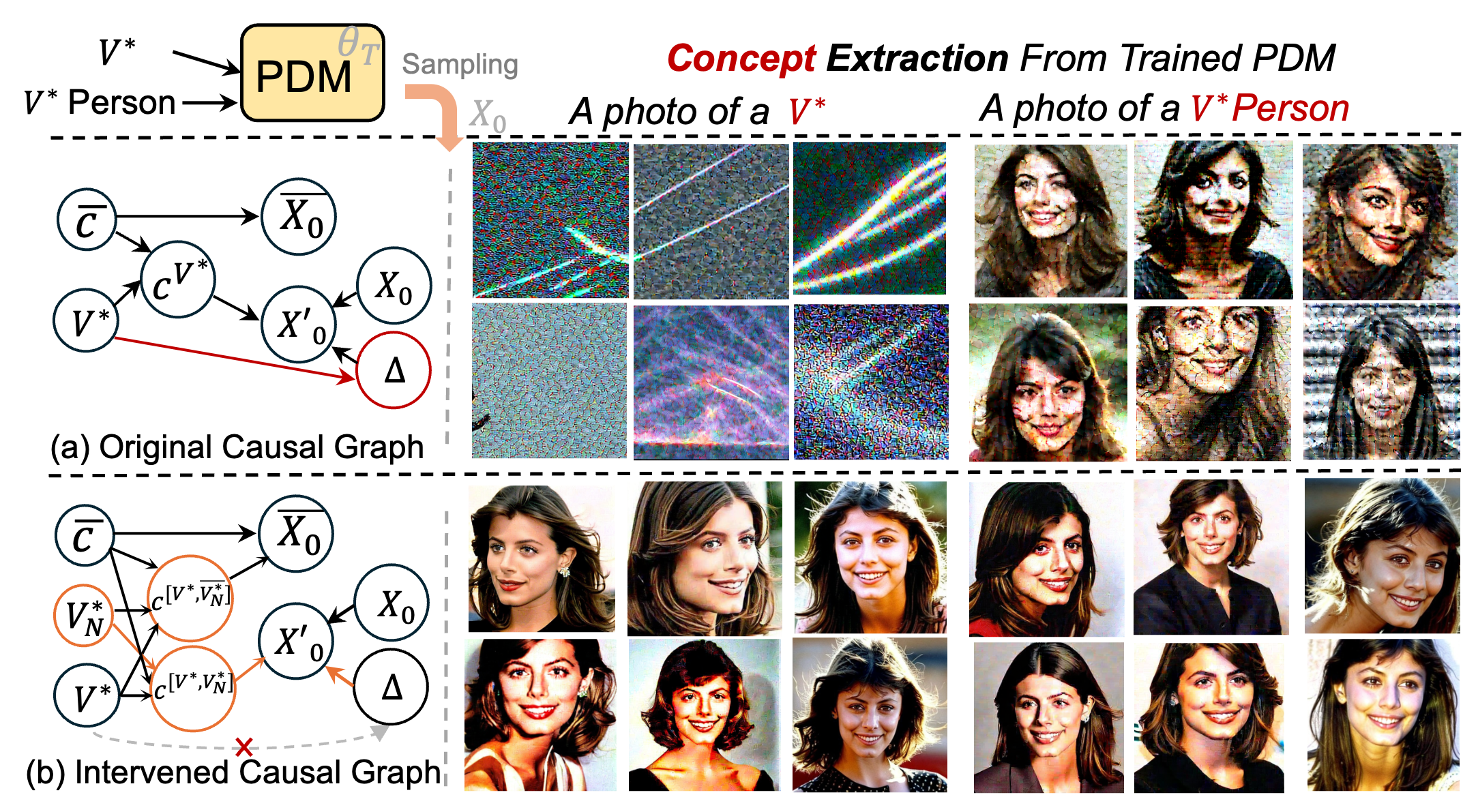}
    \caption{Visualization of concept learning with and without CDL. (a) The original causal graph where $\mathcal{V}^*$ is linked to the perturbed image $X_0'$. Prompting for ``a photo of a $\mathcal{V}^*$ Person'' incorrectly generates a noisy image. (b) With CDL, the model correctly learns the association between $\mathcal{V}^*$ and the personalized concept, generating a clean image.}
    \label{fig:causal-graph-noise}
\end{figure}

\header{Generation Visualization and Sensitivity Test.} We further visualize the generation of models trained in three cases, including clean, perturbed, and purified, in Fig.~\ref{fig:gen-vis}. The visualization demonstrates that our defense greatly helps retain clean-level generation quality.
Additionally, we find that the concept learned associated with $\mathcal{V}*$ under the perturbed case matches the noise concept learned using CDL alone, as shown in Fig.~\ref{fig:causal-graph-noise}, indicating the CDL successfully decouples the learning of noise patterns.

% Furthermore, during the sampling stage, we apply classifier-free guidance (CFG) to further improve the quality of the generated images. It modifies the generation process by incorporating negative prompts during inference, thereby adjusting the output. We define a generation function $g'(\theta_T, c^{[\mathcal{V}^*, \bar{\mathcal{V}}_N^*]}, c_{\text{neg}})$, where $g'$ is the modified generation function, $\bar{\mathcal{V}}_N^*$ denotes using ``without $\mathcal{V}_N^*$'' in the prompt, and $c_{\text{neg}}$ represents negative prompts (e.g., ``noisy, abstract, pattern, low quality''). We guide the model to generate images that do not contain any noisy pattern associated with $\mathcal{V}_N^*$ in the prompt input. This step acts as an intervention on the generation mechanism, reducing the influence of any residual associations between $\Delta$ and the outputs. Although it is more of a practical adjustment than a formal causal intervention, it helps steer the model toward generating high-quality images that reflect the personalized concept.

\header{CDL with Different Noise Tokens.}
To investigate the effect of using our CDL with different noise tokens, we present results in Tab.~\ref{table:noise_prompts}.
Our empirical evaluation shows that the choice of noise tokens can influence performance.
While common semantic descriptors (e.g., \textit{``visual interference''}) often fail to fully decouple the noise, we observe that rare or abstract tokens tend to act as more effective anchors for the specific adversarial patterns.
Among the various combinations we explored, the rare token \textit{``t@j noisy pattern''} yielded the best overall performance, successfully isolating the noise without confounding the subject's identity.
% Future works can explore automatic noise prompt searching to further optimize this selection.

\begin{table}[t]
    \centering
    \caption{Representative comparison of CDL with different noise tokens. Higher scores indicate better performance.}
    \label{table:noise_prompts}
    \resizebox{\linewidth}{!}{
        \begin{tabular}{c|ccc|cccc}
            \toprule
            \textbf{Noise Tokens} $\mathcal{V}_N*$ & \textbf{IMS} $\uparrow$ & \textbf{Q} $\uparrow$ & \textbf{Avg.} $\uparrow$ & $\text{IMS}_{\text{VGG}}$ $\uparrow$ & $\text{IMS}_{\text{IP}}$ $\uparrow$ & \textbf{LIQE} $\uparrow$ & \textbf{CLIP-IQAC} $\uparrow$ \\
            \midrule
            % Adjusted t@j to match Table 1 MetaCloak results (~0.21 IMS, ~0.47 Q)
            \textit{t@j noisy pattern} & \textbf{0.212} & \textbf{0.475} & \textbf{0.344} & 0.185 & 0.224 & \textbf{0.520} & 0.430 \\
            \textit{xjy image imperfection} & -0.131 & -0.130 & -0.130 & -0.151 & -0.122 & -0.140 & -0.120 \\
            \textit{xjy visual interference} & -0.339 & 0.014 & -0.163 & -0.413 & -0.307 & 0.028 & 0.001 \\
            \midrule
            \textit{bhi noisy perturbation} & -0.494 & -0.727 & -0.610 & -0.369 & -0.547 & -0.646 & -0.808 \\
            \bottomrule
        \end{tabular}
    }
\end{table}

\section{Discussion: Intervening on Shortcut Learning}
\label{sec:causal_intervention}
Our red-teaming strategies can be interpreted as interventions that modify the causal graph to weaken or eliminate the undesired shortcut connection between $\mathcal{V}^*$ and the noisy concept $\Delta$.

\textbf{Input Purification} aims to mitigate the effect of adversarial perturbations, but it is important to note that this process is not perfect. In the context of our causal model, we can represent this imperfect purification as:
$X_0' = X_0 + \Delta \rightarrow X_0' = X_0 + \Delta_r$,
where $\Delta_r$ represents the residual perturbations after purification, with $\|\Delta_r\| \ll \|\Delta\|$. This intervention partially weakens the path from $\Delta$ to $X_0'$, and consequently to $\theta_{T}$ and $X_0$. While input purification reduces the influence of adversarial perturbations on the fine-tuning process and subsequent image generation, it does not completely eliminate the shortcut learning problem. This limitation motivates the need for additional strategies.

\textbf{Contrastive Decoupling Learning (CDL)} intervenes on the potential shortcut $\mathcal{V}^* \rightarrow \Delta$ by introducing a noise identifier $\mathcal{V}_N^*$. By augmenting the instance prompts to include a noise identifier (e.g., ``a photo of $\mathcal{V}^*$ with $\mathcal{V}_N^*$ noisy pattern'') and augmenting the class prompts to exclude it (e.g., ``a photo of a person without $\mathcal{V}_N^*$ noisy pattern''), CDL encourages the model to disentangle the learning of the personalized concept from the noise patterns. Specifically, the model learns two clearer associations: $\mathcal{V}_N^* \rightarrow \Delta$ and $\mathcal{V}^* \rightarrow X_0$. From the results in Fig.~\ref{fig:causal-graph-noise}, we see that the decoupling process enables the model to learn two concepts separately, including the personalized concept and the noisy pattern. 

In summary, by combining these strategies, we provide a comprehensive approach to mitigate shortcut learning in PDMs. Input purification directly removes the influence of adversarial perturbations, and our CDL further reduces potential left-over spurious associations during training.

% Our red-teaming strategies function as interventions on the causal graph (as illustrated in Fig.~\ref{fig:causal-graph-noise}), designed to weaken or eliminate the undesired shortcut connection between the identifier $\mathcal{V}^*$ and the noisy concept $\Delta$. \textit{Input Purification} acts as a direct intervention on the input node, transforming the perturbed state into $X_0' = X_0 + \Delta_r$, where the residual noise satisfies $\|\Delta_r\| \ll \|\Delta\|$. This significantly weakens the causal path from $\Delta$ to the fine-tuned parameters $\theta_{T}$ and the generated image $X_0$. However, as purification is imperfect, residual associations may remain. To address this, \textit{Contrastive Decoupling Learning (CDL)} structurally intervenes on the potential shortcut $\mathcal{V}^* \rightarrow \Delta$ by introducing a dedicated noise identifier $\mathcal{V}_N^*$. By adjusting prompts to explicitly distinguish between the subject and the noise, CDL encourages the model to disentangle the representations into two independent causal mechanisms: $\mathcal{V}_N^* \rightarrow \Delta$ (noise) and $\mathcal{V}^* \rightarrow X_0$ (content). As shown in Fig.~\ref{fig:causal-graph-noise}(b), this intervention enables the model to correctly associate the identifier with the clean personalized concept rather than the noisy patterns. Together, these strategies provide a comprehensive defense by reducing input noise and structurally decoupling any remaining spurious correlations.
% % \vspace{-5pt}
% \vspace{-10pt}

\begin{table}[t]
    \centering
    \caption{Ablation study on individual modules.}
    \label{tab:ablation}
    \resizebox{.7\linewidth}{!}{
    \begin{tabular}{cccccc} 
    \toprule
    \multicolumn{3}{c}{\textbf{Settings}} & \multicolumn{3}{c}{\textbf{Metrics}}                    \\ 
    \cmidrule(lr){1-3}\cmidrule(lr){4-6}
                            \textbf{CodeF.} & \textbf{SR} & \textbf{CDL}     & \textbf{IMS} $\uparrow$ & \textbf{Q} $\uparrow$ & \textbf{Avg.} $\uparrow$  \\ 
    \midrule
    \cmarkcolor{} & \cmarkcolor{} & \cmarkcolor{} & {0.256} & \textbf{0.514} & \textbf{0.385} \\
    \cmarkcolor{} & \cmarkcolor{} & \xmarkcolor{} & -0.215 & 0.028 & -0.094 \\ 
    \cmarkcolor{} & \xmarkcolor{} & \cmarkcolor{} & \textbf{0.294} & 0.385 & 0.339 \\
    \xmarkcolor{} & \cmarkcolor{} & \cmarkcolor{} & 0.190 & 0.260 & 0.225 \\
    \cmarkcolor{} & \xmarkcolor{} & \xmarkcolor{} & -0.336 & 0.020 & -0.158 \\
    \xmarkcolor{} & \cmarkcolor{} & \xmarkcolor{} & -0.059 & -0.439 & -0.249 \\
    \xmarkcolor{} & \xmarkcolor{} & \cmarkcolor{} & 0.160 & 0.038 & 0.099 \\
    \xmarkcolor{} & \xmarkcolor{} & \xmarkcolor{} & -0.271 & -0.425 & -0.348 \\
    \bottomrule
    \end{tabular}
    }
\end{table}

\section{Conclusion}
\label{sec:conclusion}
In this work, we provide a comprehensive mechanistic diagnosis of protective perturbations designed to prevent unauthorized fine-tuning of personalized diffusion models. Through systematic analysis, we identify latent space misalignment as the primary vulnerability exploited by these perturbations, which creates spurious correlations between noise patterns and identity tokens during fine-tuning. Based on this understanding, we propose a unified red-teaming framework that addresses this challenge from two complementary perspectives: (1) an efficient purification pipeline integrating super-resolution and image restoration modules for latent realignment, which recovers the original semantic content while removing adversarial artifacts; and (2) Contrastive Decoupling Learning (CDL), a novel training strategy that explicitly separates the noise concept from the target identity to mitigate shortcut learning. Extensive experiments demonstrate that our framework achieves superior effectiveness in bypassing existing protections while maintaining high efficiency and faithfulness to the original identity.

Looking ahead, several promising directions emerge. First, our analysis (Sec.~\ref{app. resilience}) reveals a robustness-effectiveness trade-off: CodeFormer excels in standard scenarios while SR demonstrates greater resilience against adaptive attacks. Future work could explore adaptive selection mechanisms that balance utility and robustness based on detected perturbation characteristics. Second, automated prompt search strategies for CDL could enhance decoupling effectiveness by discovering more discriminative prefixes. Third, while primarily evaluated on facial data, our principles of latent realignment and contrastive decoupling are domain-agnostic and could extend to artwork protection and object-centric generation. Finally, our mechanistic insights into how protections fail can inform the design of next-generation protections that are more resilient to purification-based bypasses.

\begin{acks}
In this work, Yixin Liu and Lichao Sun were partially supported by the National Science Foundation Grants CRII-2246067, ATD-2427915, NSF POSE-2346158, NSF POSE-2449280, and Lehigh Grant FRGS00011497. This work is also supported by the Delta/DeltaAI systems at NCSA and the Bridges-2 system at PSC through allocation CIS240308 from the Advanced Cyberinfrastructure Coordination Ecosystem: Services \& Support (ACCESS) program, supported by National Science Foundation grants 2138259, 2138286, 2138307, 2137603, and 2138296.
\end{acks}

\clearpage
\bibliographystyle{ACM-Reference-Format}
\bibliography{ref}

\appendix
\section{Implementation Details}
\label{app. exp-details}

\subsection{Metrics}
\label{app. metric}
In this section, we describe the evaluation metrics used in our experiments in more detail. Following~\citep{Liu_2024_CVPR}, we use CLIP-IQAC, which calculates the CLIP score difference between ``a good photo of [class]'' and ``a bad photo of [class]''. For calculating IMS-VGGNet, we leverage the VGGNet in the DeepFace library for face recognition and face embedding extraction \citep{serengil2021lightface}. For IMS-IP, we leverage \textit{antelopev2} model from InsightFace library \citep{Deng2020CVPR} following IP-adapter \citep{ye2023ip-adapter}. We report the weighted average of them with a weighting factor on IMS-IP as 70\% since we find it yields a more stable evaluation with IMS-VGG as 30\%.
We compute all the mean scores for all generated images and instances. For the instance $i$ and its $j$-th metric, its $k$-th observation value is defined as $m_{i,j,k}$. For the $j$-th metric, the mean value is obtained with $\sum_{i,k}m_{i,j,k}/(N_i N_k)$, where $N_i$ is the instance number for that particular dataset, and $N_k$ is the image generation number.

\subsection{Hardware and Training Details}
\label{app. training details}
\header{Hardware Details.} All the experiments are conducted on an Ubuntu 20.04.6 LTS (focal) environment with 503GB RAM, 10 GPUs (NVIDIA\textsuperscript{\textregistered} RTX\textsuperscript{\textregistered} A5000 24GB), and 64 CPU cores (Intel\textsuperscript{\textregistered} Xeon\textsuperscript{\textregistered} Silver 4314 CPU @ 2.40GHz). Python 3.9.18 and PyTorch 1.13.1 are used for all the implementations. For our proposed purification pipeline, the peak VRAM usage is 1625MB for the CodeFormer component and 4129MB for the Super-Resolution component per batch of 4 samples. The entire fine-tuning and inference process for a single subject can be completed on a single 24GB GPU.

\header{Training and Inference Settings. }The Stable Diffusion (SD) v2-1-base \citep{Rombach_2022_CVPR} is used as the model backbone. For DreamBooth training, we conduct full fine-tuning, which includes both the text-encoder and U-Net model with a constant learning rate of $5 \times 10^{-7}$ and batch size of 2 for 1000 iterations in mixed-precision training mode. We use the 8-bit Adam optimizer with $\beta_1 = 0.9$ and $\beta_2 = 0.999$ under bfloat16-mixed precision and enable the xformers for memory-efficient training. For calculating prior loss, we use 200 images generated from Stable Diffusion v2-1-base with the class prompt \texttt{``a photo of a [class noun]''}. The weight for prior loss is set to 1.
For the evaluation phase, we set the inference steps as 100 with prompts ``a photo of sks person'' and ``a smiling photo of sks person'' during inference to generate \textit{16} images per prompt. For all the settings, the classifier-free guidance \cite{ho2022classifier} is turned on by default with a guidance scale of $7.5$. This value is adopted following common practice in the classifier-free guidance literature and our empirical observations that it yields good generation quality. For Contrastive Decoupling Learning, we use ``\texttt{t@j}'' as the noise token $\mathcal{V}_N^*$.

\subsection{CodeSR Purification Configuration}
\label{app:sr-code-config}
Table~\ref{tab:sr_code_config} presents the detailed configuration parameters for our CodeSR purification pipeline. These parameters were carefully selected to balance generation quality and fidelity preservation.

\begin{table}[ht]
\centering
\resizebox{\linewidth}{!}{
\begin{tabular}{lll}
\toprule
\textbf{Parameter} & \textbf{Value} & \textbf{Description} \\
\midrule
Input/Output Resolution & 512$\times$512 & Image resolution \\
SR Model & Stable Diffusion x4 Upscaler & Super-resolution model \\
SR Intermediate Size & 128$\times$128 & Downscaled size before SR \\
CodeFormer Weight (w) & 0.5 & Balance quality vs fidelity \\
Noise Level & 20 & Noise level for diffusion \\
% Guidance Scale & 9.0 & CFG strength for SR upscaler \\
\bottomrule
\end{tabular}
}
\caption{CodeSR Purification Configuration Parameters}
\label{tab:sr_code_config}
\end{table}

\section{Causal Analysis of Learning Personalized Diffusion Models on Perturbed Data}
\label{appendix:causal_analysis}

% \subsection{Construction of the Causal Graph when Learning PDMs on Perturbed Data}
% \label{appendix:causal_graph}
To understand how \blueUpdate{protective} perturbations lead to shortcut learning in PDMs, we construct a Structural Causal Model (SCM) that captures the \blueUpdate{learned} causal relationships between the variables involved in the fine-tuning process. The variables in our SCM are defined as follows: $\blueUpdate{X_0}$ represents the original clean images representing the true concept; $\blueUpdate{\Delta}$ denotes the protective perturbations added to the images; $\blueUpdate{X_0^\prime = X_0 + \Delta}$ are the perturbed images used for fine-tuning; $c$ represents class-specific textual prompts without the unique identifier (e.g., ``a photo of a person''); $\mathcal{V}^*$ is the unique identifier token used in personalized prompts (e.g., ``sks''); $c^{\mathcal{V}^*} = c \oplus \mathcal{V}^*$ denotes the personalized textual prompts combining $c$ and $\mathcal{V}^*$; $\theta_{\blueUpdate{T}}$ represents the model parameters after being fine-tuned. The structural equations governing the relationships in our SCM are as follows: (1) Perturbed Images: $X_0' = X_0 + \Delta$, where $X_0'$ represents the perturbed images, $X_0$ the original clean images, and $\Delta$ the protective perturbations. (2) Model Fine-tuning: $\theta_{\blueUpdate{T}} = f_{\theta}(\theta_0, X_0', c^{\mathcal{V}^*}, \bar{X_0} ,\bar{c})$, where $\theta_{\blueUpdate{T}}$ represents the fine-tuned model parameters, $\theta_0$ the initial model parameters, $c^{\mathcal{V}^*}$ the personalized text prompts, $\bar{X_0}$ and $\bar{c}$ the image and prompt of class-specific dataset to help model maintain class prior. For our case of fine-tuning on human portrait, the $\bar{X_0}$ is the person images from different identities, and $\bar{c}$ is set as ``a photo of a person''. After $\theta_T$ has been fine-tuned, it learns the latent causal relationship $\mathcal{V}^*$ $\rightarrow$ $X_0'$ with conditioning mechanism through prompt-image association.

Based on these equations, we construct a causal graph shown in Fig.~\ref{fig:causal-graph-noise} (a) following the conventions in causal inference. The main causal graph is presented in Fig.~\ref{fig:causal-graph-noise} in the main paper.
In the graph, we define each node to represent one of the elements for the learned causation: independent variables (i.e., text prompts, and unique identifier), dependent variables (i.e., perturbed identity images, general face images), or intermediate variables like prompt composited. We define each edge to represent the causal unidirectional dependency between the variables. For those prompt composition edges, the relationship is simply the concatenation operation in the textual space. For those prompt-image association edges, the relationship is defined as the causation learned by the model $\theta_T$. For the edges between $\Delta$ and $X_0'$, it is defined as the direct effect of the perturbations on the original clean images, $X_0'= X_0 + \Delta$. Similar to the confounder in causal inference, we can see from the graph that the perturbation $\Delta$ induces a shortcut connection from the unique identifier $\mathcal{V}^*$ to the noisy concept $\Delta$. Note that in the context of backdoor learning through causal inference, such confounder is termed trigger or backdoor variable \citep{zhang2023backdoor,pmlr-v235-liu24bu}. Different from the backdoor scenario, our case of protective perturbation introduces confounding variables on the learning target side instead of on the input side in backdoor attacks.

\section{Ethical Statement}
\label{sec:ethical}
This work red-teams adversarial protections to assess real-world robustness. While our methods could potentially bypass privacy controls, this research advances the field by revealing that current protections suffer from shortcut learning vulnerabilities, providing critical insights for designing next-generation defenses. All experiments were conducted using publicly available datasets with appropriate usage licenses. Beyond this specific application, the core idea of contrastive decoupling learning can be extended to other generative AI security scenarios where mitigating shortcut learning and separating desired concepts from spurious patterns are essential. We support responsible disclosure and encourage developers to prioritize resilience against adaptive purification in future protection designs.

\end{document}